\documentclass[journal]{IEEEtran}
\usepackage{amsmath,amsfonts}
\usepackage{algorithm, algorithmic}
\usepackage{array}
\usepackage{textcomp}
\usepackage{stfloats}
\usepackage{url}
\usepackage{verbatim}
\usepackage{graphicx}
\usepackage[table,xcdraw]{xcolor}
\usepackage{booktabs}
\usepackage{multirow}
\usepackage{cite}
\usepackage{mathtools, cuted}
\usepackage{lipsum, color}
\usepackage{hyperref}
\hypersetup{
    colorlinks=true,
    linkcolor=blue,
    filecolor=magenta,      
    urlcolor=cyan,
    pdftitle={Overleaf Example},
    pdfpagemode=FullScreen,
    }
\usepackage{balance}
\ifCLASSINFOpdf

\else

\fi

\begin{document}

\title{\emph{Tombo} Propeller: Bio-Inspired Deformable Structure toward Collision-Accommodated Control for Drones}

\author{Son~Tien~Bui$^{1)}$,
    Quan~Khanh~Luu$^{1)}$,
    Dinh~Quang~Nguyen$^{1)}$,
    Nhat~Dinh~Minh~Le$^{1)}$,\\
    Giuseppe~Loianno$^{2)}$,
    and Van~Anh~Ho$^{1)3)*}$% <-this % stops a space
\thanks{$^{1)}$Soft Haptics Lab., School of Materials Science, Japan Advanced Institute of Science and Technology, Asahidai 1-1, Nomi, Ishikawa, 923-1292 Japan}% <-this % stops a space
\thanks{$^{2)}$Tandon School of Engineering, New York University, 11201 Brooklyn, NY 11201, USA.}% <-this % stops a space
\thanks{$^{3)}$Japan Science and Technology Agency, PRESTO, Kawaguchi Saitama 332-0012 Japan.}
\thanks{Data, implemented code for this paper has been updated at: \url{https://github.com/Ho-lab-jaist/tombo-propeller.git}}
\thanks{$^{*}$Corresponding author. Email: {\tt\small  van-ho@jaist.ac.jp}}}% <-this % stops a space

% The paper headers
\markboth{This work has been submitted to the IEEE for possible publication. Copyright may be transferred without notice, after which this version may no longer be accessible}%
{Shell \MakeLowercase{\textit{et al.}}: }

% make the title area
\maketitle

\begin{abstract}
There is a growing need for vertical take-off and landing vehicles, including drones, which are safe to use and can adapt to collisions. The risks of damage by collision, to humans, obstacles in the environment, and drones themselves, are significant. This has prompted a search into nature for a highly resilient structure that can inform a design of propellers to reduce those risks and enhance safety. Inspired by the flexibility and resilience of dragonfly wings, we propose a novel design for a biomimetic drone propeller called Tombo propeller. Here, we report on the design and fabrication process of this biomimetic propeller that can accommodate collisions and recover quickly, while maintaining sufficient thrust force to hover and fly. We describe the development of an aerodynamic model and  experiments conducted to investigate performance characteristics for various configurations of the propeller morphology, and related properties, such as generated thrust force, thrust force deviation, collision force, recovery time, lift-to-drag ratio, and noise. Finally, we design and showcase a control strategy for a drone equipped with Tombo propellers that collides in mid-air with an obstacle and recovers from collision continuing flying. The results show that the maximum collision force generated by the proposed Tombo propeller is less than two-thirds that of a traditional rigid propeller, which suggests the concrete possibility to employ deformable propellers for drones flying in a cluttered environment. This research can contribute to morphological design of flying vehicles for agile and resilient performance.
\end{abstract}

\begin{IEEEkeywords}
deformable propeller, collision-accommodated, biomimetic design, soft robotics, drones' safety
\end{IEEEkeywords}

\IEEEpeerreviewmaketitle

\section{Introduction} \label{sec: introduction}
\IEEEPARstart{D}{rones} such as popular VTOL (vertical take-off and landing) have brought enormous benefits to humans operating in various sectors, providing military, civilian, and commercial applications, including monitoring, inspection, logistics, transportation, and entertainment. Due to their compact dimension and flight agility in small spaces, drones have been attracting interest from both academia and industry with the potential of large market \cite{UAVmarket}. Several projects were launched to harness the potential of drones in industry, such as Prime Air (Amazon - 2013), Wing (Alphabet 2014), DRONES (FedEx - 2018), FarmBeats (Microsoft in partnership with DJI - 2015), Aquila (Facebook - 2014), Skylink (IBM - 2016), and others. One of the primary concerns in the operation of drones is the risk of injury to humans or damage to property-both objects in the environment and the drone itself, which may occur during flight or in event of a collision. In such a scenario, a drone could seriously harm a human or significantly damage something and drop and crash due to damaged propellers. With that in mind, many technologies promoting the safety of drones have been introduced, such as safety cages for drones (hardware), and vision-based obstacle-avoiding algorithms (intelligence). While the former increases the size and physical load of a drone, the latter increases the computational load on its central processor, resulting in the trade off of high efficiency costs for safety. Nevertheless, the desire to operate drones in cluttered environments or nearby humans is growing and necessitates strategies to mitigate injuries or damages caused by unexpected collisions.
\begin{figure}
	\centering
	\includegraphics[width=\linewidth]{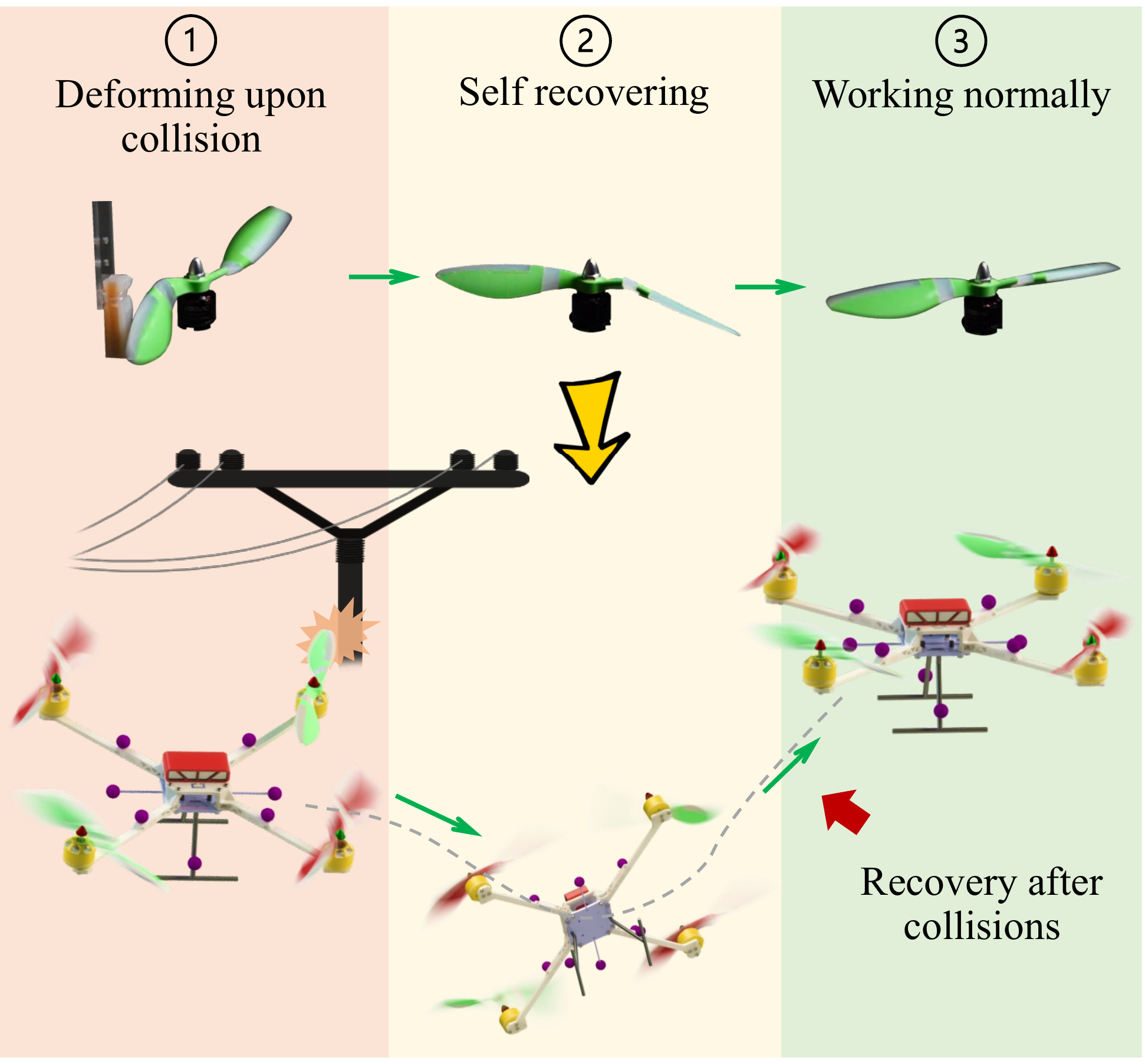}
	\caption{Safety drone equipped with Tombo propellers. Collision-accommodated propellers that can deform passively upon collision then self-recover to work normally are employed for uncrashed drones.}
	\label{fig1}
\end{figure}

\par Meanwhile, in the natural world, flying insects seem to effortlessly accommodate collisions with objects in their surroundings without leaving any footprint of damage. In fact, mother nature is the greatest creator, and at the same time she is a living encyclopedia from which robotics researchers may learn invaluable knowledge and glean hints for solving engineering problems. In this paper, the dragonfly wing construction is an excellent example of shock absorption and self-recovery, due to the high flexibility of a hinge-like structure called \textit{nodus}. From the perspective of soft robotics, biomimicry of such structures or functions is a key to solve the safety problem of drones.

\par This paper presents the Tombo propeller. Without requiring the burden of a heavy physical or computational load, toward ultimate safety for drones, this biomimetic design deformable propeller can accommodate collisions while retaining sufficient rotation-thrust to remain airborne as shown in Figure \ref{fig1}. The key structure of the Tombo propeller is its hinge-like \textit{nodus}, which is a deformable joint made of silicone and fiber tendons (see Figure \ref{fig: design of deformable propeller}a) bio-inspired by the function and structure of the dragonfly wing nodus. Due to the  \textit{nodus} mechanism, a Tombo propeller can self-recover and rotate properly within an average of 0.46\,seconds upon collision, enabling the promising recovery of drones after sudden collisions with their surroundings. In addition, the hybrid design made of soft and hard parts enables the propeller to regain stiffness and generate sufficient rotation-thrust to fly the drone. Moreover, with its deformable leading edge, the Tombo propeller imparts lower impact force and damage upon collision with an object than that by a traditional drone propeller, thus further improving the safety of the propeller and struck object. The main contributions of this research are as follows:
\begin{enumerate}
    \item Proposal of a novel biomimetic design for drone propellers. %(section \ref{sec: design})
    \item Construction of an aerodynamic model of the Tombo propeller. %(section \ref{sec: aerodynamicsmodel}), 
    \item Theoretical and experimental characterization of the Tombo propeller with different configurations of \textit{nodus}. %(section \ref{sec: characteristicsTombo}),
    \item Proposal of a control strategy which includes collision recovery by a drone with Tombo propellers.
\end{enumerate}
Compared to our previous paper \cite{foldableprop}, where we only introduced design, fabrication, and preliminary measurement of thrust force of the Tombo propeller; in this paper, we present a thorough analysis and evaluation of the propeller considering variations in its material's composition and configuration as well as different rotational speeds.  In addition, we propose a model for characterization of generated thrush force, taking into account the deformation of the Tombo propeller and its aerodynamic properties. We propose several metrics to evaluate and identify the optimal design according to the user needs. Finally, we also test the flight ability of real drone platform using the Tombo propeller, and proposed a simplified control strategy for showing the feasibility of recovery of drone after the collision.    
\par In this paper, a concrete review of drones and related work is mentioned in Section \ref{sec: relatedworks}, followed by the biomimetic design of a 9-inch (228.6\,mm) long Tombo propeller in Section \ref{sec: design}. Section \ref{sec: aerodynamicsmodel} briefly introduces a derivation of an aerodynamics model for the Tombo propeller. Section \ref{sec: characteristicsTombo} provides details of Tombo propeller characteristic measurement experiments, which is followed by the results in Section \ref{sec: results}. Next, Section \ref{sec: flying experiment} presents a control strategy and showcases the flying of an embedded Tombo propeller drone. Then, is the Discussion in Section \ref{sec: discussion} and Conclusion in Section \ref{sec: conclusion}.

\section{Related Works} \label{sec: relatedworks}
In this section, we focus on solutions that have been introduced to improve the safety of drones and Unmanned Aerial Vehicles (UAVs). Overall, these solutions can be classified into two groups: \textit{Collision Avoidance} and \textit{Collision Impact Reduction}. Details can be found below:

For collision avoidance, the following approaches have been proposed: geometric, force-field, optimized, sense and avoid \cite{avoidancetypes, avoidancetypes2}. Here, passive sensors (monocular camera \cite{monocularcamera}, depth camera \cite{depthcamera}), active sensors (ultrasonic sensor \cite{ultrasonicsensor}, light detection and ranging \cite{2D-LiDarsensor}, and radar \cite{radarsensor}) or a combination of them \cite{RGB-Dsensor, RGB-Dsensor2} are preferable to detect objects within the flying range. Although such cameras have a large field of view and high resolution, their associated  obstacle detection algorithm is greatly affected by weather conditions, lighting, and shiny or reflective surfaces. In addition, vision sensors work best with stationary subjects, yet they are required to passively respond to high-speed moving images exceeding $35$ km/h \cite{35kmh} and are limited at blind spots. Moreover, sensors are needed to overcome the blind spot factor and improve the accuracy of calculation, resulting in increased weight and higher cost UAVs. Recently, event cameras have been applied to UAVs with promising results in terms of avoiding fast-moving obstacles \cite{eventcamera}. However, the event camera has a heavier weight, larger size, and higher noise than a standard one at the same resolution. Furthermore, the high noise of an event camera reduces its accuracy at large distances, limiting the reliable detection range to $1.5$\,m.

\par Regarding collision impact reduction, solutions to protect the actuators (rotors) from collisions, or facilitate recovery after a collision require additional features. In most cases, a cage \cite{eulerspring,protectivecages, protectivecages2, collisionresilientrobot} is employed due to low cost, low profile, and simple setup; for instance, protective cages shielding motors and preventing multi-directional concurrences. However, their bulky structure may inversely bring a higher risk of collision, and they greatly increase the drone weight thus reducing the flight time. As an alternative approach, foldable structures \cite{foldabledrone1, foldabledrone2} could help UAVs decrease their dimension and risk of crashing, especially when flying through a narrow gap. Another exciting approach combines rigid guards and soft, deformable mechanisms to reduce crash impact, absorb collision shock, and retain resilience  \cite{resilientmulticopters, foldablearm,rotorigami, impactresilient}. However, such integration may require a complex control strategy and high resource consumption in practical scenarios. Recently, the notion of a flexible blade \cite{flexibleblade} has gained attention since this structure may enable propellers to deform and recover upon collision without any additional mechanism. Nonetheless, a flexible blade remains dangerous for soft objects such as human skin. Furthermore, large-sized thin blades experience great deformation during rotation, limiting the use of these propellers in large UAVs.

\par Consequently, a trade-off between the safety of UAVs and their structure and configuration exists. Therefore, despite efforts to improve the structure, integrate perception, and avoid collisions, drones and UAVs remain vulnerable to collisions because of technical limitations of sensors and unpredictable factors. This poses a great challenge, as nowadays, UAVs are becoming increasingly popular and being used for more application in complex, cluttered, or partially-known environments. Hence, there remains a big question: Is there a solution that significantly improves the safety of drones but does not largely compromise their basic design and function?

\section{Biomimetic Design} \label{sec: design}
\begin{figure*}[!t]
	\centering
		\includegraphics[width=\textwidth]{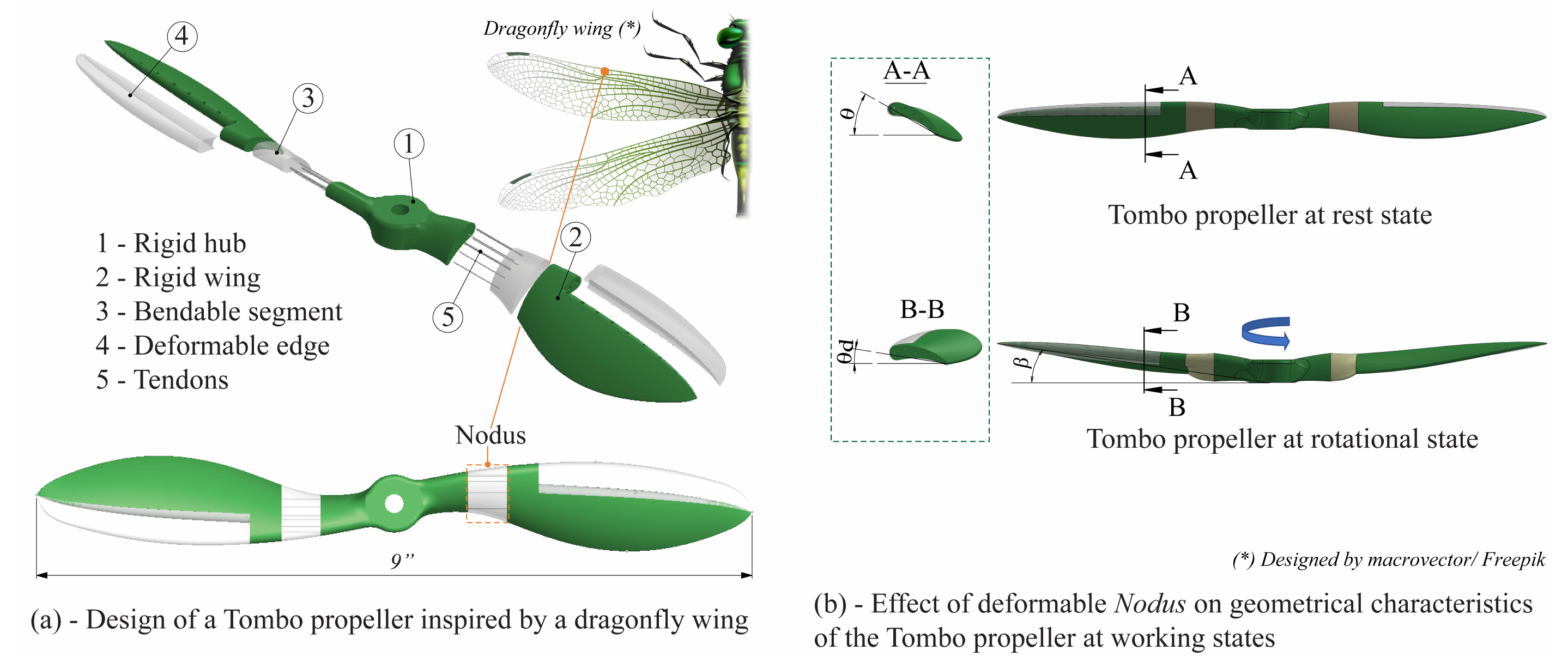}
	\caption{Tombo propeller anatomy: (a) The design of a Tombo propeller includes a rigid hub, two wings, two soft blades, and two \textit{nodus} with tendons inside. (b) The effect of the deformable \textit{nodus} on the geometrical parameters of the propeller results in decreasing thrust generation. In return, due to \textit{nodus} flexibility, the Tombo propeller can self-recover and work normally upon collision.}
	\label{fig: design of deformable propeller}
\end{figure*}
\par Insect wings often suffer stress due to wind gusts and collisions with the surroundings. Therefore, the insect wings evolved to adopt an anti-shock or shock-absorption structure, resulting in mitigation of wing damage upon collision. Regarding the dragonfly wing structure, one of the most important parts is the \textit{nodus}, which has a one-way hinge-like structure that can passively flex upon external contact. This part allows the wing to twist, bend, or even fold without incurring damage. The natural \textit{nodus} containing resilin (similar to isotropic rubber) can provide both structural reinforcement and shock absorption. These characteristics provide a hybrid structure (stiff and flexible) that both ensure aerodynamic properties when flying and reduce the risk of damage to the wing upon collision. Based on the characteristics of the dragonfly wing nodus, we propose an unprecedented deformable propeller which we named Tombo propeller (Figure \ref{fig: design of deformable propeller}a). The rigid parts include a hub (1) and wing (2), connected by a nodus-like bendable segment (3) made from silicon rubber and reinforced by embedded tendons (5, nylon fibers) that connect rigid parts (hub and wings). An optional deformable edge (4) can be added for rapid absorption of an impact at the time of collision. In our initial work \cite{foldableprop}, the rigid parts of the propellers were made of ABS (Acrylonitrile Butadiene Styrene) using FFF (fused filament fabrication)-based 3D printing which creates a layered structure, which eventually reduces the strength among layers upon lateral impact. Therefore, in this work we propose a novel generation model, the rigid parts are fabricated by injection molding as commercial rigid propellers to enhance the concreteness of the Tombo propeller. To enhance the propeller surface quality, especially the \textit{nodus}-like part, molds were made of aluminum alloy fabricated by machining instead of ABS fabricated by 3D printing. The fabrication process for a Tombo propeller has been described \cite{foldableprop}.

\par \textit{Nodus} mechanical property and performance strongly affects the working ability of the proposed propeller. As illustrated in Figure \ref{fig: design of deformable propeller}, when the propeller rotates, there is a decrease in pitch angle theta, which is due to the inherent propeller's deformation. This negatively affects thrust force generation since it depends largely on the pitch angle. In addition, the softness of the \textit{nodus} may influence the working stability of a drone equipped with Tombo propellers. Therefore, in this research, we investigate various configurations of the \textit{nodus} structure and materials (silicone rubber embedded with monofilament nylon fibers), \emph{i.e.} the morphology, to determine an optimum construction of the Tombo propeller.

\section{Aerodynamic model of Tombo propeller} \label{sec: aerodynamicsmodel}
Previous research has indicated that aerodynamic parameters such as drag and thrust forces \cite{aerodynamicseffects1, aerodynamicseffects2, aerodynamicseffects3}, blade flapping \cite{aerodynamicseffects2, bladeflapping1}, and so on, strongly affect the flight characteristics of drones. Control approaches for drones require consideration to aerodynamic parameters to attain stability during dynamic flight or hovering \cite{Haddadin}. Therefore, in developing an aerodynamics model of the Tombo propeller, mechanical properties play an important role for its application to standard UAVs. 

\subsection{Revisit of Aerodynamic Model of a Classical Propeller}

\par Aerodynamic models of standard propellers and quadrotors have been established and well developed \cite{MarinePropellersandPropulsion, windturbine, marineprop}. The most important factors of aerodynamic parameters, the forces, including the  normal force ($F_n$), tangential force ($F_t$), lift force ($F_l$), and drag force ($F_d$) (see Figure \ref{fig: nodusmodelling}e) can be calculated as
\begin{align}
	{F_i} &= \int_{{P}} \rho \omega^2  {C_i} \mathrm{d}{S} x^2 \mathrm{d}x,\hspace{0.6cm} i \in \left\{ n, t, l, d \right\} 
	\label{eq: 1}
\end{align}
where ${P}$ is the designed span length function along the airfoils; $\rho$ is the air density; $\omega$ is rotational speed; ${C_i} = {C_i}({\theta}) $ is the aerodynamics force coefficient function (see \cite{aerodynamic_coefficient}) and ${\theta} = {\theta}(x)$ is the designed pitch angle function (Figure \ref{fig: nodusmodelling}e); ${S}$ is the designed boundary surface function; $x$ is the span length element along the airfoils.

\subsection{The Role of Deformable Angles}
\begin{figure*}
	\centering
    \includegraphics[width=0.98\textwidth]{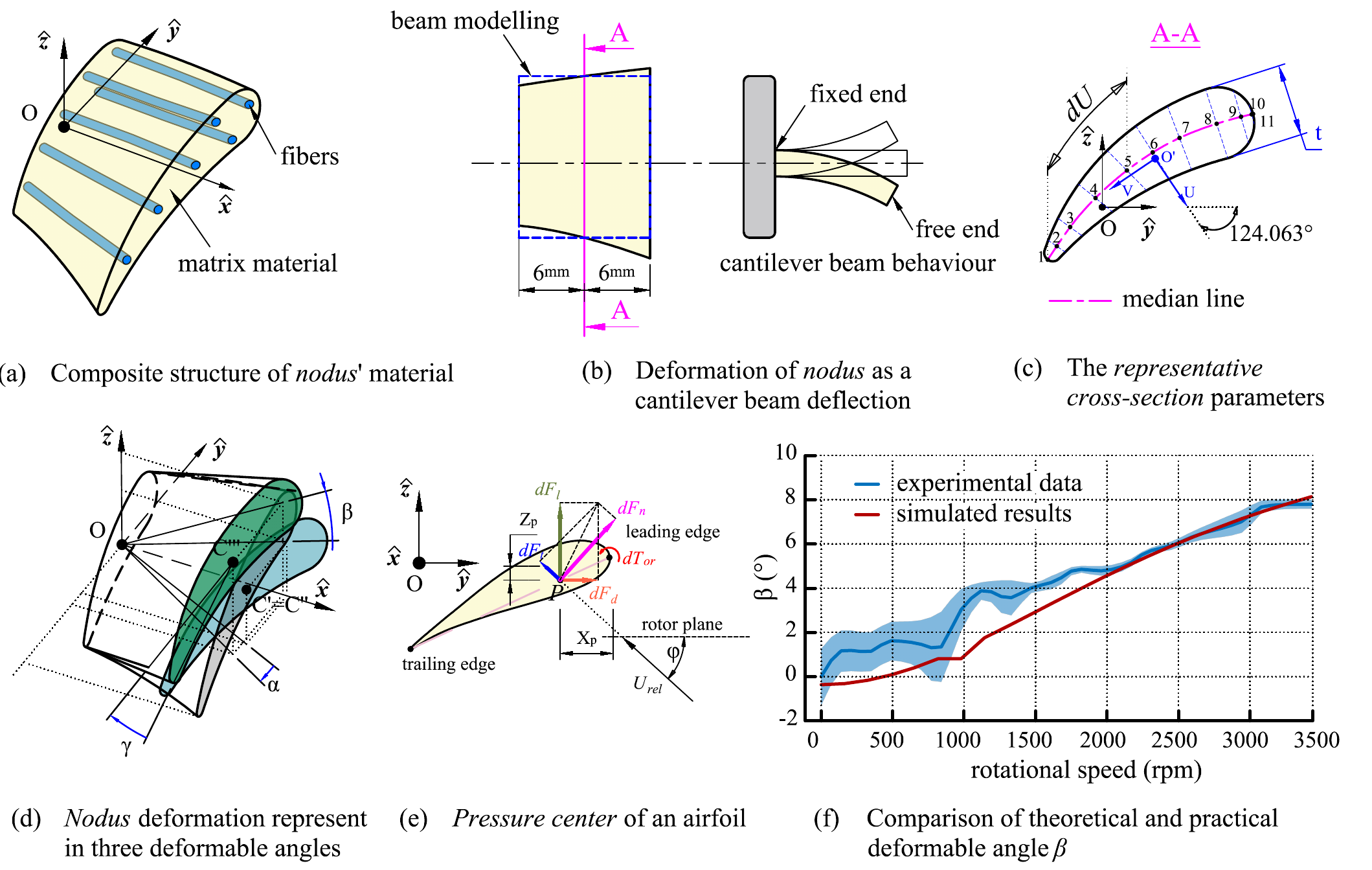}
	\caption{\textit{Nodus} modelling and evaluation of deformable angle $\beta$. (a) The body of the \textit{nodus} can be described as a composite structure comprising reinforced fibers in a material matrix. (b) A cantilever beam with a cross-section as the \textit{representative cross-section} stay at the center of the \textit{nodus} length enables deformable behavior of the nodus. (c) Parameters of the \textit{representative cross-section} include the key median line (or the mean camber line) interpolated by the ten point coordinates shown. (d) The deformable behavior of the \textit{nodus}: 1) from  $C'$ to $C''$: rotated around $OC''$ by $\gamma$, 2) from $C''$ to $C'''$: rotated around $Oz$ and $Oy$ by $\alpha$ and $\beta$ in turn. Note that $O$ and $C'$ are the center points at either end of the nodus. (e) The \textit{center of pressure} stays at the quarter-chord of an airfoil. $U_{rel}$ is the relative wind. (f) The simulated deformable angle $\beta$ and the experimental with reference to the rotational speed.}
	\label{fig: nodusmodelling}
\end{figure*}

Although the flexibility of the \textit{nodus} saves the propeller from collision damage, it creates unexpected deformation when the propeller rotates. In detail, when a rotor is rotating, rigid wing (Figure \ref{fig: nodusmodelling}d) displacement is quantitatively specified by bending angles $\alpha$, $\beta$ and twist angle $\gamma$ as depicted in Figure \ref{fig: nodusmodelling}d. These deformable angles alter the geometry of the propeller, such as pitch angle $\theta$, inducing change in aerodynamic forces $F_d$ and $F_l$. Therefore, an aerodynamic model Tombo propeller needs to factor in the deformable angles of the \textit{nodus} in order to achieve efficient operation of the device.

\subsection{\textit{Nodus} Modelling} \label{IV-C}

\par To determine the deformable angles, we model both material and structure, \emph{i.e.} morphology, of the \textit{nodus} using the combination of composite model and beam model.
\subsubsection{Material Modelling}
The structure of the \textit{nodus} suggests a composite model of silicone rubber and a tendon play the roles of material matrix and reinforced fibers, respectively (see Figure \ref{fig: nodusmodelling}a). Based on research by Younes \emph{et al.} on different models for composite modelling \cite{Compositereview}, we chose the Chamis model for modelling because of its high accuracy for predicting elastic modulus coefficients in many matrix and fiber material cases. According to the Chamis model, Young modulus $E_N$ and the shear modulus $G_N$ of \textit{nodus} are defined as below:
\begin{equation}
    E_N= \frac{E^m}{1-\sqrt{v^f}(1-\frac{E^m}{E{f}})}, 
    G_N=\frac{G^m}{1-\sqrt{v^f}(1-\frac{G^m}{G^{f}})},
    \label{eq: 2}
\end{equation}
where $v^f$ is the fiber volume fraction; $E^m$ and $G^m$ are the Young modulus and shear modulus of the matrix material, respectively; $E^{f}$ and $G^{f}$ are the Young modulus and shear modulus of the fiber material in turn. The fiber volume fraction $v^f$ (from $0$ to $1$), measured as the percentage of fiber area in the composite cross-section, dominates the mechanical properties of the \textit{nodus}. Changing the number of fibers or their diameter adjusts fiber volume fraction $v^f$, resulting in a proportional change in \textit{nodus} stiffness.

\subsubsection{Structure Modelling}
We hypothesized that the \textit{nodus} deflects obeying cantilever beam behavior with one fixed end (see Figure \ref{fig: nodusmodelling}b) during its rotation. Therefore, the deformable angles can be defined in a model of a stationary cantilever (at a specific rotation speed) by applying the beam modelling. As a result, bending angles $\alpha$ in $Oxy$ plane and $\beta$ in $Oxz$ plane of a Tombo propeller are defined based on \cite{cantileverbeammodelling} as follows:
\begin{align}
	\alpha=\dfrac{F_{BN}^y L_N^2}{2E_N I_N^z} \hspace{1cm}
	\beta=\dfrac{F_{BN}^z L_N^2}{2 E_N I_N^y},
	\label{eq:alpha and beta}	
\end{align}
where $F_{BN}^y$, $F_{BN}^z$ are forces from the wing applied to the \textit{nodus}; $L_N$ is the \textit{nodus} length; $E_N$ is the Young modulus of the \textit{nodus}; $I_N^y$ and $I_N^z$ are the inertial moments of the cross-section of the \textit{nodus} in $\hat{y}$, $\hat{z}$ directions, respectively. Since the inertial moments depend on the cross-section position, Eq. \ref{eq:alpha and beta} shows a strong relationship between \textit{nodus} length and the cross-section position to deformable angles $\alpha$ and $\beta$.
\subsubsection{Hybrid Modelling}
  To model the mechanical characteristics of the \textit{nodus}, we used hybrid modelling combining insights of both aforementioned material and structural models. However, both popular composite models reviewed in \cite{Compositereview} and the beam model \cite{cantileverbeammodelling} often required high homogeneity of materials and a standard cross-section area structure; while the \textit{nodus} has different directions of nylon tendons and varying morphology along its body. Therefore, for ease of modeling, we proposed a \textit{representative cross-section} (Figure \ref{fig: nodusmodelling}c), stay at the middle of the \textit{nodus} (section A-A) to overcome this issue. This particular cross-section will be used for all related calculations of the \textit{nodus}. The twist angle $\gamma$ of a Tombo propeller can be defined as a form of elongated cross-section as below \cite{elongatedcrosssection}:
\begin{align}
	\gamma=\dfrac{3(1+\dfrac{4F}{3A_N U^2})T_{or} L_N}{G_{N}F},
	\label{eq: gamma}	
\end{align}
where $F=\int_0^Ut^3\mathrm{d}U$, $\mathrm{d}U$ is  the infinitesimal length along the camberline line; $T_{or}$ is the applied torque generated by the aerodynamic forces $F_d$ and $F_t$; $A_N$ and $U$ are the area and the length of a camberline of the \textit{representative cross-section}; $L_N$ are the length of the \textit{nodus}; $t$ is the thickness normal to the median line. If we call P is \textit{the center of pressure} at \textit{the quarter-chord point} of the airfoil \cite{windturbine, Anderson2013} (see Figure \ref{fig: nodusmodelling}e). The applied torque element $\mathrm{d}T_{or}$ can be defines as below.
%\begin{align}
%    y_P=\frac{\mid y_{LE}-y_{TE} \mid}{4};
%    z_P=\frac{\mid z_{LE}-z_{TE} \mid}{4}
%\end{align}
%Then
\begin{align}
    \mathrm{d}T_{or}=\mathrm{d}F_{t} \frac{\mid y_{LE}-y_{TE} \mid}{4} +  \mathrm{d}F_{d} \frac{\mid z_{LE}-z_{TE} \mid}{4}
\end{align}
with $y_{LE}$, $z_{LE}$, $y_{TE}$, and $z_{TE}$ are the coordinates of leading edge and trailing edge of the \textit{representative cross-section} in $\hat{x}$ and $\hat{z}$ directions respectively.
\subsection{Aerodynamic Model of Tombo Propeller} \label{subsec: aeodynamicsmodel}
At a rotational speed, we assumed the deformable propeller morphology was unchanged. Therefore, the applied aerodynamic forces of a Tombo propeller can be calculated as in Eq. \ref{eq: 1}. Note that the optional deformable edge (4) of the Tombo propeller was assumed not to deform in our proposed aerodynamic model at rotational speed. Thus, the influence of the deformable edge is negligible in the computational model. Here, functions of the span length, aerodynamic force coefficient, and the boundary surface were considered to be the main contributors to the deformable angles. In the model, the aerodynamic forces consist of three components generated by the hub, the \textit{nodus} so these aerodynamic forces can be explained as below:
\begin{equation}
	{F_i} = {F_i^{h}} + {F_i^{N}} + {F_i^{w}}, \hspace{1cm} i \in \left\{ n, t, l, d \right\}
	\label{eq: 6}
\end{equation}
The hub is rigid and the \textit{nodus} length is about  10\,$\%$ of that of the propeller so we can use Eq. \ref{eq: 1} to calculate both ${F_i^{h}}$ and ${F_i^{N}}$. The aerodynamic force of the wing \cite{windturbine} for the Tombo propeller needs to take into account the contribution of the deformable angles as below.
% \begin{align}
% 	{F_i} &= \int_{\boldsymbol{\Re}{.C^d}} \rho \omega^2  {C_i^d} \mathrm{d}(\boldsymbol{\Re}{.S^d}) x^2 \mathrm{d}x,\hspace{0.6cm} i \in \left\{ n, t, l, d \right\}
% 	\label{eq: 5}
% \end{align}
% where $\boldsymbol{\Re}\textbf{.}$ is the skew-symmetric matrix operator such that $\boldsymbol{\Re}\textbf{.}a = \boldsymbol{R}a$, $\boldsymbol{R}$ is the transpose matrix function of the \textit{representative cross-section}; ${C_i^d} = {C_i({\theta^d})}$ with ${\theta^d}$ is the deformable pitch angle function of the Tombo propeller airfoil; ${C^d}= \boldsymbol{\Re}\textbf{.}{C}$ and ${S^d}=\boldsymbol{\Re}\textbf{.}{S}$ are the deformable span length function along the airfoils and the deformable boundary surface function of the Tombo propeller. In the model, the aerodynamic forces consist of three components generated by the hub, the \textit{nodus}, and the wing is as follows:
\begin{align}
	{F_i^w} &= \int_{{P^{de}}} \rho \omega^2  {C_i^{de}} \mathrm{d}({S^{de}}) x^2 \mathrm{d}x,\hspace{0.6cm} i \in \left\{ n, t, l, d \right\}
	\label{eq: 5}
\end{align}where ${P^{de}}$ is the deformable span length function along the airfoils; ${C_i^{de}} = {C_i({\theta^{de}(x)})}$ with ${\theta^{de}}$(x) is the deformable pitch angle function of the Tombo propeller airfoil; ${S^{de}}$ is the deformable boundary surface function; ${P^{de}}$ and ${S^{de}}$ can be defined by the projections of the wing into the $\hat{x}$ direction and $Oxy$ plane respectively when \textit{nodus} deforms. Note that the wing posture can be defined by rotation matrix $\mathbf{R}(\gamma,\beta, \alpha) \in SO(3)$ using a roll-pitch-yaw sequence of rotations around axes of a fixed reference frame \cite{rotmat}:
\begin{equation}
    \mathbf{R}(\alpha, \beta, \gamma) = \mathbf{R}(\hat{z}, \alpha)\mathbf{R}(\hat{y},\beta)\mathbf{R}(\hat{x},\gamma)
\end{equation}
where ${\alpha}$, ${\beta}$ are the bending angle functions of the \textit{nodus} on $Oxz$, $Oxy$; ${\gamma}$ is the twist angle around the centroid contour along the \textit{nodus}; $\mathbf{R}(\hat{x},\gamma)$, $\mathbf{R}(\hat{y},\beta)$, and $\mathbf{R}(\hat{z}, \alpha)$ are the rotations around the $\hat{x}$, $\hat{y}$, and $\hat{z}$ axes of the fixed reference frame (see Figure \ref{fig: nodusmodelling}d), respectively.

% The hub is rigid and the \textit{nodus} length is about  10\,$\%$ of that of the propeller so we can use Eq. \ref{eq: 1} to calculate both ${F_i^{h}}$ and ${F_i^{N}}$. Rotation matrix function $\mathbf{R}$ for wing deformation can be defined by rotation matrix function $\mathbf{R_1}$ around $Oz$ \& $Ox$ and rotation matrix function $\mathbf{R_2}$ around the normal vector of the \textit{representative cross-section} after rotating:

To simplify change in boundary surface function, we project this wing into three original planes $Oxy$, $Oyz$, and $Ozx$. Finally, we obtained the equation of aerodynamic forces applied to a wing as below:
\begin{multline}
	{F^w_i} = \int_{r_N}^{R} \rho \omega^2 {C_i^{de}} \dfrac{\cos{\alpha}\cos {\beta}\cos ({\theta} - {\gamma})}{\cos{{\theta}}} \left({L} - {T}\right)    x^2 \mathrm{d}x\\ i \in \left\{ n, t, l, d \right\},
	\label{eq: 8}
\end{multline}
where $r_N$ and $R$ are radius of innermost and outermost curvatures from the position of \textit{nodus} onward; ${L}$, ${T}$ are the designed geometrical functions of the leading edge and the trailing edge, respectively.
\begin{figure}
	\centering
    \includegraphics[width=0.9\columnwidth]{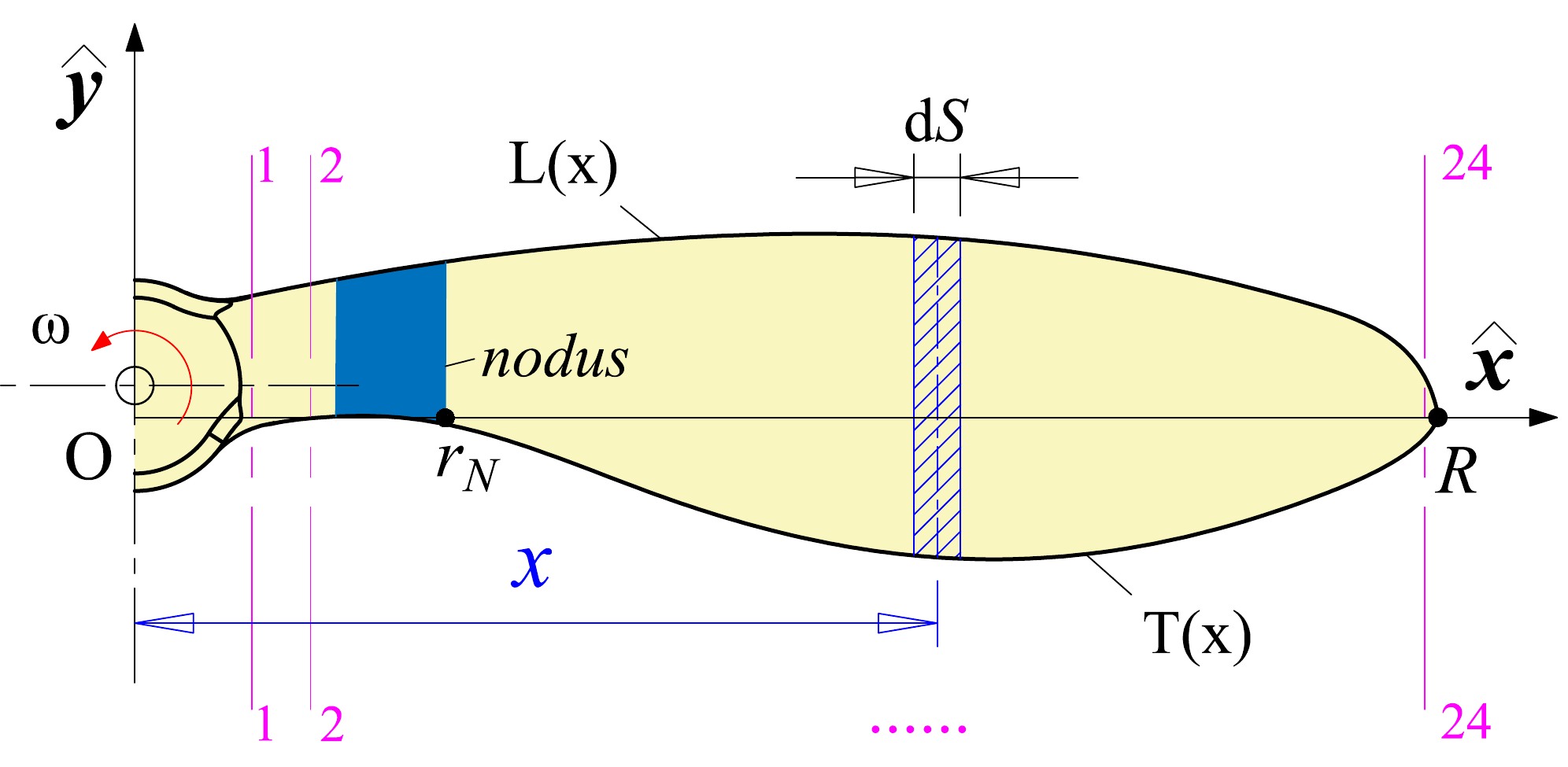}
	\caption{A half of designed propeller shows the boundary surface element $\mathrm{d}({S})$ and the geometrical functions ${L}$ and ${T}$. Magenta lines locate the positions of cross-sections for collecting airfoil information.}
	\label{fig: halfprop}
\end{figure}
\par To define functions ${L}$, ${T}$ we represent half of a propeller by 24 cross-sections along the span (Figure \ref{fig: halfprop})  by SolidWorks 2020 software. In each Section we collected the angle of attack and the coordinates of leading edges and trailing edges by AutoCad 2022 software. Next, we applied the formula of ${L}$, ${T}$, and ${\theta}$ as functions of span length using \textit{polyfit} function of Matlab R2020b.
\subsection{Numerical Implementation of the Aerodynamic Model}
By its nature, the proposed model has an implicit problem because of the inter-dependencies between the input parameters. Non-linear solution techniques can solve this problem, but using a simple algorithm where an iteration loop is used until convergence is reached is a more practical solution. Algorithms 1 and 2 have shown this approach in two cases: a classical propeller (the rigid one) and a Tombo propeller (see \url{https://github.com/Ho-lab-jaist/tombo-propeller.git} for more details).
% Algorithm 1
% The result can be used to confirm the equation of C_L and C_D
\begin{algorithm}
	\caption{Aerodynamics Parameters Estimation for Rigid Propellers}
	\begin{algorithmic}[1]
		\renewcommand{\algorithmicrequire}{\textbf{Input:}}
		\renewcommand{\algorithmicensure}{\textbf{Output:}}
		\REQUIRE Airfoil parameters, air density $\rho$, and rotational speed of propeller $n_r$
		\STATE $N$ := number of cross-sections
		\STATE ${S_C} = \{S_i: [x_{C_i}, y_{LE_i}, z_{LE_i}, y_{TE_i}, z_{TE_i}, \theta_{C_i}] \}$ := coordinate of leading edge and trailing edge, and pitch angle $i \in \{ 1,...,N \}$ \\
		\ENSURE  Arodynamics forces $F_j$, $j \in \left\{ n, t, l, d \right\}$ \\
		\STATE //\textit{Separate geometric functions into parts by order}
		\STATE $k$ := Number of geometrical function parts
		\STATE $o_i$ :=  Order of the $i$th part
		\STATE $F_i := 0$
		%\STATE $D_{RP} := 0$
		\FOR {$i := 1$ to $k$}
		\STATE $P_i :=$ polyfit$(x_C,y_{LE},y_{TE}, o_i)$
		\STATE $F_j:=$ polyfit$(x_C,{C_j}(\theta_C),o_i)$
		%\STATE $D_i:=$ polyfit$(x_C,\boldsymbol{C_d}(\theta_C),o_i)$
		%\STATE $T_{RP} = T_{RP} + \boldsymbol{T_{RP}}(\rho, n_r, P_i, L_i)$
		%\STATE $D_{RP} = D_{RP} + \boldsymbol{D_{RP}}(\rho, n_r, P_i, D_i)$
		\ENDFOR
		\RETURN $F_j$ 
	\end{algorithmic} 
\end{algorithm}

\makeatletter
\newcommand{\ALOOP}[1]{\ALC@it\algorithmicloop\ #1%
	\begin{ALC@loop}}
	\newcommand{\ENDALOOP}{\end{ALC@loop}\ALC@it\algorithmicendloop}
\renewcommand{\algorithmicrequire}{\textbf{Input:}}
\renewcommand{\algorithmicensure}{\textbf{Output:}}
\newcommand{\algorithmicbreak}{\textbf{break}}
\newcommand{\BREAK}{\STATE \algorithmicbreak}
\makeatother
% Algorithm 2
\begin{algorithm}
	\caption{Aerodynamics Parameters Estimation for Tombo Propellers}
	\begin{algorithmic}[1]
% 		\show\LOOP
		\renewcommand{\algorithmicrequire}{\textbf{Input:}}
		\renewcommand{\algorithmicensure}{\textbf{Output:}}
		\REQUIRE Airfoil functions, representative section, \textit{nodus} parameters, air density $\rho$, and rotational speed of propeller $n_r$, loop options
		\STATE ${S_C}$, $k$, $P_i$, $L$, $T$ := Airfoil parameter and functions
		\STATE $U_C$, $A_C$, $F$ := \textit{Representative cross-section} parameters
		\STATE $E_{N}, G_{N}, I^x_{C}, I^z_{C}, x_N, L_N$ := Young's and shear modulus, coordinate and length of \textit{nodus}
		\STATE $nbIt$ := Maximum number of iterations
		\STATE $Tol$ := Acceptance tolerance
		\ENSURE  Arodynamics parameters $F_j$, $C_j$, $j \in \left\{ n, t, l, d \right\}$, $Tor$, lift-over-drag ratio $\varepsilon_{lod}$
		\STATE //LOOP
		%\show\LOOP {$i := 1$ to $nbIt$}
		\ALOOP {} % Outer loop
		\STATE $\alpha = {\alpha}(F_{d}, E_{N}, I^z_{C})$ 
		\STATE $\beta = {\beta}(F_{t}, E_{N}, I^x_{C})$
		\STATE $T_{or} = {T_{or}}(F_{t2}, F_{d2}, \theta_C) $
		\STATE $\gamma = \boldsymbol{\gamma}(T_{or}, G_{N}, F_C)$
		\STATE $F^w_{j} = {F_{j}}(\rho, n_r, \theta_C, C_j, x_N, L_N)$, $j \in \left\{ n, t, l, d \right\}$
		%\STATE $D_{DP2} = \boldsymbol{D_{DP}}(\rho, n_r, \theta_C, D_i, x_N, L_N)$
		\STATE //\textit{Calculate the results}
		\STATE $F_{j} = F^h_{j} + F^N_{j} + F^w_{j}$
		\STATE //Convergence Criteria
		\IF{$i>nbIt$}
		\BREAK
		\ELSE
		\IF{$\left|F_t - F_t{last}\right| + \left|F_d-F_{dlast}\right| < Tol$}
		\BREAK
		\ENDIF
		\ENDIF
		% \ELSIF{$i := nbIt}
		\ENDALOOP
		\RETURN $F_j$, $C_j$, $T_{or}$, and $\varepsilon_{lod}$
	\end{algorithmic} 
\end{algorithm}

\section{Measurement of Tombo Propeller's Characteristics } \label{sec: characteristicsTombo}

\begin{figure*}[!t]
	\centering
    \includegraphics[width=\textwidth]{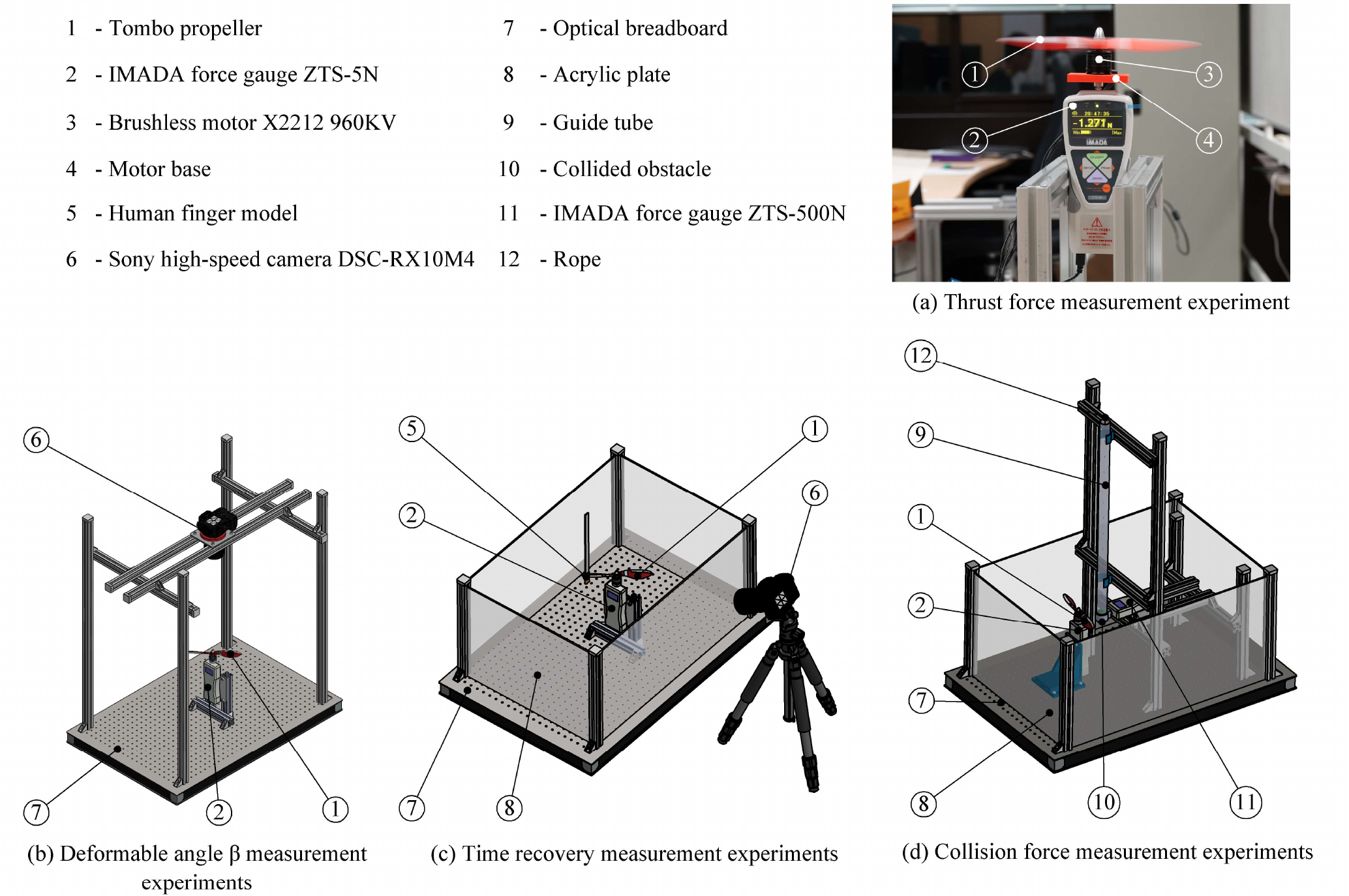}
    \caption{The measurement setup for investigating the Tombo propeller characteristics. (a) Thrust force measurement experiments. (b) Deformable angle $\beta$ measurement experiments. (c) Time recovery measurement experiments. (d) Collision force measurement experiments.}
	\label{fig: 4experiments}
\end{figure*}
\par In this section, we describe indoor experiments with different configurations of \textit{nodus} matrix materials and fiber diameters (see Table \ref{tab:Tombo configuration}) to determine the characteristics of Tombo propellers. These experiments were performed for the following purposes:
\begin{enumerate}
    \item To evaluate the aerodynamic model proposed in \ref{sec: aerodynamicsmodel} based on the thrust force and deformation angle results.
    \item To observe characteristics of the Tombo propeller in action to acquire and build fundamental knowledge on this biomimetic propeller.
\end{enumerate}
\begin{table*}
\caption{Tombo propeller configurations used to evaluate the aerodynamic model and observe working characteristics}
\label{tab:Tombo configuration}
\begin{tabular}{|l|llrrrrr|l|r|}
\hline
\multicolumn{1}{|c|}{\multirow{2}{*}{Name}} &
  \multicolumn{7}{c|}{\textit{Nodus}' configuration} &
  \multicolumn{1}{c|}{\multirow{2}{*}{\begin{tabular}[c]{@{}c@{}}Deformable edge\\materials\end{tabular}}} &
  \multicolumn{1}{c|}{\multirow{2}{*}{\begin{tabular}[c]{@{}c@{}}Size \\ (inch)\end{tabular}}} \\ \cline{2-8}
\multicolumn{1}{|c|}{} &
  \multicolumn{1}{c|}{Matrix material} &
  \multicolumn{1}{c|}{\begin{tabular}[c]{@{}c@{}}Fiber\\ material\end{tabular}} &
  \multicolumn{1}{c|}{\begin{tabular}[c]{@{}c@{}}Number of\\ fibers\end{tabular}} &
  \multicolumn{1}{c|}{\begin{tabular}[c]{@{}c@{}}Diameter of\\ fiber (mm)\end{tabular}} &
  \multicolumn{1}{c|}{\begin{tabular}[c]{@{}c@{}}Length\\ (mm)\end{tabular}} &
  \multicolumn{1}{c|}{\begin{tabular}[c]{@{}c@{}}Young's modulus\\ $E_N$ (MPa)\end{tabular}} &
  \multicolumn{1}{c|}{\begin{tabular}[c]{@{}c@{}}Shear modulus\\ $G_N$ (MPa)\end{tabular}} &
  \multicolumn{1}{c|}{} &
  \multicolumn{1}{c|}{} \\ \hline
Conf. 0 &
  \multicolumn{1}{c|}{-} &
  \multicolumn{1}{c|}{-} &
  \multicolumn{1}{c|}{-} &
  \multicolumn{1}{c|}{-} &
  \multicolumn{1}{c|}{-} &
  \multicolumn{1}{c|}{-} &
  \multicolumn{1}{c|}{-} &
  \multicolumn{1}{c|}{-} &
  9 \\ \hline
Conf. 1 &
  \multicolumn{1}{l|}{DragonSkin 10} &
  \multicolumn{1}{l|}{Nylon 6} &
  \multicolumn{1}{r|}{6} &
  \multicolumn{1}{r|}{0.94} &
  \multicolumn{1}{r|}{12} &
  \multicolumn{1}{r|}{0.1542} &
  0.0551 &
  \multicolumn{1}{c|}{-} &
  9 \\ \hline
Conf. 2 &
  \multicolumn{1}{l|}{DragonSkin 20} &
  \multicolumn{1}{l|}{Nylon 6} &
  \multicolumn{1}{r|}{6} &
  \multicolumn{1}{r|}{0.94} &
  \multicolumn{1}{r|}{12} &
  \multicolumn{1}{r|}{0.5705} &
  0.2036 &
  \multicolumn{1}{c|}{-} &
  9 \\ \hline
Conf. 3 &
  \multicolumn{1}{l|}{DragonSkin 30} &
  \multicolumn{1}{l|}{Nylon 6} &
  \multicolumn{1}{r|}{6} &
  \multicolumn{1}{r|}{0.94} &
  \multicolumn{1}{r|}{12} &
  \multicolumn{1}{r|}{0.7873} &
  0.2809 &
  \multicolumn{1}{c|}{-} &
  9 \\ \hline
Conf. 4 &
  \multicolumn{1}{l|}{DragonSkin 30} &
  \multicolumn{1}{l|}{Nylon 6} &
  \multicolumn{1}{r|}{6} &
  \multicolumn{1}{r|}{0.38} &
  \multicolumn{1}{r|}{12} &
  \multicolumn{1}{r|}{0.6557} &
  0.2341 &
  \multicolumn{1}{c|}{-} &
  9 \\ \hline
Conf. 5 &
  \multicolumn{1}{l|}{DragonSkin 10} &
  \multicolumn{1}{l|}{Nylon 6} &
  \multicolumn{1}{r|}{5} &
  \multicolumn{1}{r|}{0.5} &
  \multicolumn{1}{r|}{12} &
  \multicolumn{1}{r|}{0.1315} &
  0.0469 &
  DragonSkin 10 &
  9 \\ \hline
Conf. 6 &
  \multicolumn{1}{l|}{DragonSkin 10} &
  \multicolumn{1}{l|}{Nylon 6} &
  \multicolumn{1}{r|}{5} &
  \multicolumn{1}{r|}{0.75} &
  \multicolumn{1}{r|}{12} &
  \multicolumn{1}{r|}{0.1413} &
  0.0505 &
  DragonSkin 10 &
  9 \\ \hline
Conf. 7 &
  \multicolumn{1}{l|}{DragonSkin 10} &
  \multicolumn{1}{l|}{Nylon 6} &
  \multicolumn{1}{r|}{5} &
  \multicolumn{1}{r|}{0.9} &
  \multicolumn{1}{r|}{12} &
  \multicolumn{1}{r|}{0.148} &
  0.0528 &
  DragonSkin 10 &
  9 \\ \hline
Conf. 8 &
  \multicolumn{1}{l|}{DragonSkin 20} &
  \multicolumn{1}{l|}{Nylon 6} &
  \multicolumn{1}{r|}{5} &
  \multicolumn{1}{r|}{0.5} &
  \multicolumn{1}{r|}{12} &
  \multicolumn{1}{r|}{0.4862} &
  0.1736 &
  DragonSkin 20 &
  9 \\ \hline
Conf. 9 &
  \multicolumn{1}{l|}{DragonSkin 20} &
  \multicolumn{1}{l|}{Nylon 6} &
  \multicolumn{1}{r|}{5} &
  \multicolumn{1}{r|}{0.75} &
  \multicolumn{1}{r|}{12} &
  \multicolumn{1}{r|}{0.5227} &
  0.1866 &
  DragonSkin 20 &
  9 \\ \hline
Conf. 10 &
  \multicolumn{1}{l|}{DragonSkin 20} &
  \multicolumn{1}{l|}{Nylon 6} &
  \multicolumn{1}{r|}{5} &
  \multicolumn{1}{r|}{0.9} &
  \multicolumn{1}{r|}{12} &
  \multicolumn{1}{r|}{0.5473} &
  0.1954 &
  DragonSkin 20 &
  9 \\ \hline
Conf. 11 &
  \multicolumn{1}{l|}{DragonSkin 30} &
  \multicolumn{1}{l|}{Nylon 6} &
  \multicolumn{1}{r|}{5} &
  \multicolumn{1}{r|}{0.5} &
  \multicolumn{1}{r|}{12} &
  \multicolumn{1}{r|}{0.671} &
  0.2396 &
  DragonSkin 30 &
  9 \\ \hline
Conf. 12 &
  \multicolumn{1}{l|}{DragonSkin 30} &
  \multicolumn{1}{l|}{Nylon 6} &
  \multicolumn{1}{r|}{5} &
  \multicolumn{1}{r|}{0.75} &
  \multicolumn{1}{r|}{12} &
  \multicolumn{1}{r|}{0.7213} &
  0.2574 &
  DragonSkin 30 &
  9 \\ \hline
Conf. 13 &
  \multicolumn{1}{l|}{DragonSkin 30} &
  \multicolumn{1}{l|}{Nylon 6} &
  \multicolumn{1}{r|}{5} &
  \multicolumn{1}{r|}{0.9} &
  \multicolumn{1}{r|}{12} &
  \multicolumn{1}{r|}{0.7552} &
  0.2695 &
  DragonSkin 30 &
  9 \\ \hline
\end{tabular}
\end{table*}
\par The experiment apparatus included: an X2212 $960$\,KV motor (to rotate the propeller) fed by a $12$\,V power source, which was mounted on a force gauge (IMADA ZTS-5N 10Hz, Japan); an acrylic base plate to isolate the conductor and other parts (section \ref{subsec: thrustforce}; a high-speed camera (DSC-RX10M4, Sony, Japan) to record recovery time and deformable angle $\beta$ measurements (sections \ref{subsec: timerecovery} and \ref{subsec: collisionforce})); lighting and a black background to facilitate visual detection. The results of the experiments are presented and discussed in Section \ref{sec: results}.
\subsection{Thrust Force Measurement Experiments} \label{subsec: thrustforce}
In this experiment, we measured the thrust force of the Tombo propeller with reference to the  rotational speed through the range $2000$ to $3200$\,rpm with $13$ configurations, as shown in Table \ref{tab:Tombo configuration}. The experiment was conducted indoors, and the axis of the motor was set vertical relative to the ground (see Figure \ref{fig: 4experiments}a). The measurement sampling rate was $10$\,Hz. 
\subsection{Deformable Angle Measurement Experiments} \label{subsec: deformableangle}
\par For this measurement, the high-speed camera ($960$\,fps) was set perpendicular to the rotor plane and coincident with the rotor axis as shown in Figure \ref{fig: 4experiments}d. The camera distance and light source intensity were adjusted to obtain good quality images.
\par
\begin{figure*}
	\centering
    \includegraphics[width=\textwidth]{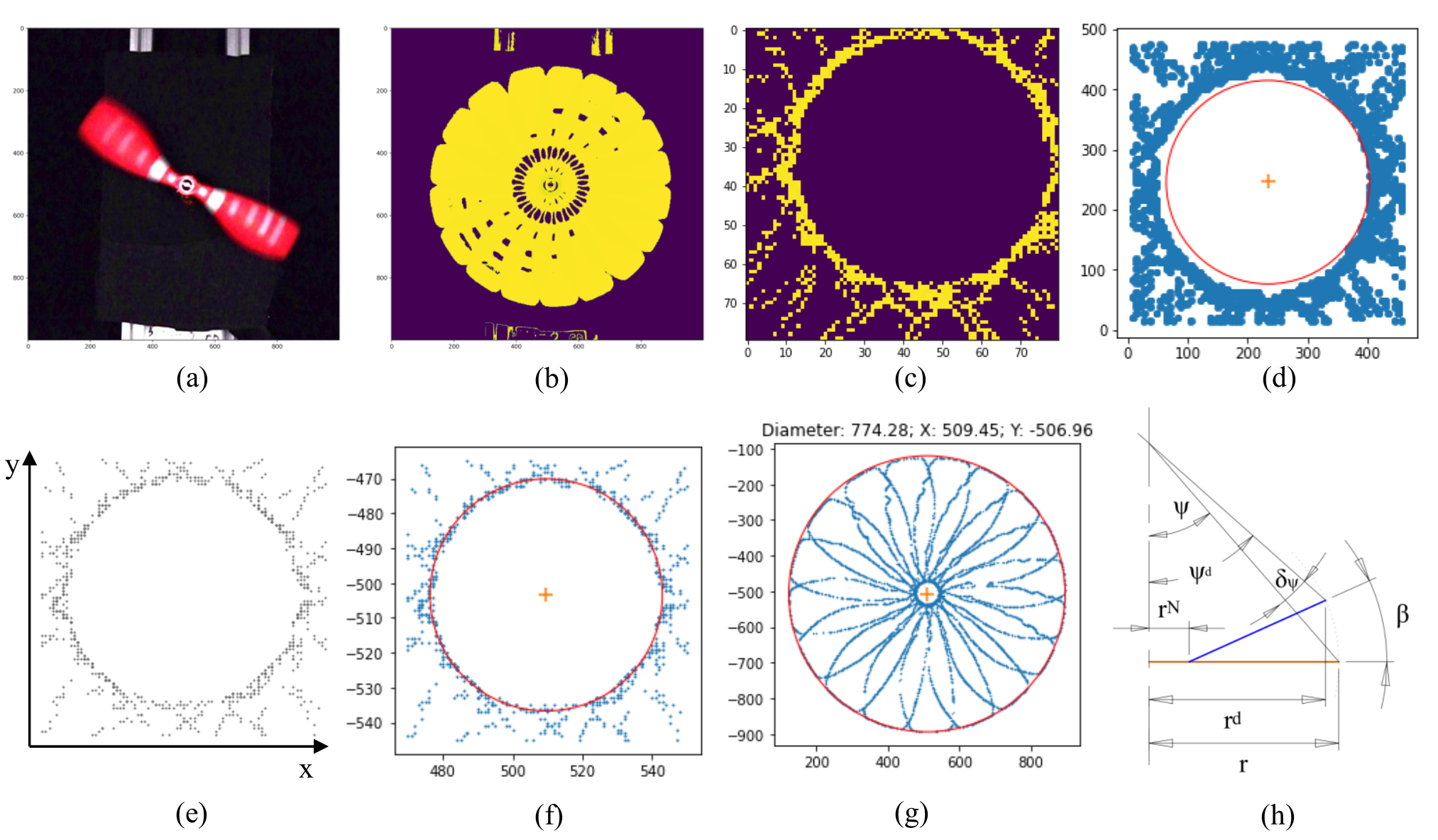}
	\caption{Method of deformation angle $\beta$ detection. (a) Images (blurred) were recorded of the high speed rotating propeller. (b) A $10$-continuous-frame combo was used for speed normalization. (c) The center of a $10$-frame combo image shows as a graph of pixels. (d) A failure of rotational center detection by Open CV deployed function. (e) In transition from image processing to numerical processing, the pixel image is represented by a real number graph. (f) Result of rotational center detection by a developed function based on the RANdom SAmple Consensus algorithm. (g) Due to the convex Hull, the propeller diameter can be rapidly determined and visualized. (h) Definition of deformable angle as a function of propeller diameter change: $\beta = acos{(\frac{r^d - r^N}{r - r^N})}$ with $r^d$ being the initial radius of a Tombo propeller, $r$ is the radius of one in a deformed state, $r^N$ is the distance from the center of the propeller to the start face of \textit{nodus}, $\psi$ and $\psi^d$ are the camera view angles in rest state and deformable state of Tombo propeller, $\delta_{\psi} =\psi^d - \psi$ is the change of camera view angles.}
	\label{fig: algo_motion_blur}
\end{figure*}

 Initially, we increased the speed of the rotor to a specific rotation speed, then we started recording and reduced the rotor speed to $0$\,rpm. The recorded video was processed by OpenCV in Python and split into a series of still image frames, which appeared blurred (Figure \ref{fig: algo_motion_blur}a). Next, we determined the rotational speed of the Tombo propeller in each image frame and the corresponding propeller diameter using a normalized 10-continuous-frame combo (Figure \ref{fig: algo_motion_blur}b).To determine the propeller diameter at each combo, we needed to focus on the center of its rotation (Figure \ref{fig: algo_motion_blur}c). Center detection initially attempted using OpenCV function was poor because the generated coordinate must be an integer (Figure \ref{fig: algo_motion_blur}d). We then transformed the coordinates of each pixel to the real field (Figure \ref{fig: algo_motion_blur}e). From that, we utilized a RANdom SAmple Consensus (RANSAC)  \cite{ransac} algorithm to determine the appropriate center of a point cloud (Figure \ref{fig: algo_motion_blur}f). We then could define the diameter of a $10$-continuous-frame combo taking advantage of Hull's contour (Figure \ref{fig: algo_motion_blur}g). Finally, the deformable angle $\beta$ could be calculated based on change in diameter and the position of the \textit{nodus} (Figure \ref{fig: algo_motion_blur}h). Note that we omitted recorded image perspective distortions made by the change of propeller tip in the vertical direction due to the small increment of the camera angle of view  $\delta_{\psi}$ (less than $1.2$$\,$$\%$). 
\subsection{Time Recovery Measurement Experiments} \label{subsec: timerecovery}
\par The objective of this experiment was to determine the recovery time of the Tombo propeller after a collision. In this experiment we used a model artificial finger made of silicone rubber (Dragon Skin 30) with a nylon mono filament core (see Figure \ref{fig: 4experiments}b) and the high-speed camera to record the process of collision and subsequent recovery of the propeller (for both rotational speed and thrust). The propeller recovery time in terms of thrust force was calculated from measurements by both the force sensor and the camera, while that in terms of rotational speed was determined from the video data.
\subsection{Collision Force Measurement Experiments} \label{subsec: collisionforce}
For consistent measurement, the direction of the collision force must coincide with the measurement axis of the force gauge. The ZTS-500N force gauge with a large force range (up to $500$\,N), and a tube ($3$\,cm in diameter, $150$\,cm in length) was placed perpendicular to the motor plane (see Figure \ref{fig: 4experiments}c). The principle of measuring the collision force was as follows. An obstacle was dropped freely inside the guide tube from the top end to collide with the propeller at a specific location, and the collision force was recorded by the force gauge.

\par This experiment was divided into two parts. In part 1, we determined the most dangerous crash zone on the propeller blade by colliding the object at different distances $50$, $60$, $70$, $80$, $90$, and $100$\,mm from the center of the propeller (see Table \ref{tab: critical collision positions}). The position at $90$\,mm from the center of the propeller was chosen to perform part 2 of the test due to its highest critical force per thickness ($61.4$\,N/mm). In part 2, we measured the collision force over a range of Tombo propeller configurations from Conf. 5 through Conf. 13 (see Table \ref{tab:Tombo configuration}). The test results are summarized in Table \ref{tab: survey of Tombo propellers}.

\begin{table*}
\caption{Collision force and thickness of collided blade of a Tombo propeller Conf. 13 at $2500$\,rpm}
\label{tab: critical collision positions}
\centering
\begin{tabular}{|l|r|r|r|r|r|r|}
\hline
\textbf{Collision area} &
  \multicolumn{1}{c|}{\textbf{I}} &
  \multicolumn{1}{c|}{\textbf{II}} &
  \multicolumn{1}{c|}{\textbf{III}} &
  \multicolumn{1}{c|}{\textbf{IV}} &
  \multicolumn{1}{c|}{\textbf{V}} &
  \multicolumn{1}{c|}{\textbf{VI}} \\ \hline
Distance to the hub center (mm) & $50$ & $60$ & $70$ & $80$ & $90$ & $100$ \\ \hline
Maximum collision force (N) & $206.5$ & $337.2$ & $331.2$ & $244.4$ & $225.3$ & $136.3$  \\ \hline
Thickness of the blade (mm)      & $6.2$   & $6.2$   & $6.0$   & $5.3$   & $3.7$   & $2.8$    \\ \hline
Force per thickness (N/mm)     & $33.2$  & $54.4$  & $55.2$  & $45.8$  & \cellcolor[HTML]{99FF66}$61.4$  & $49.1$   \\ \hline
\end{tabular}
\end{table*}
\subsection{Noise Measurement Experiments} \label{subsec: noise}
Here, the experiment employed a handheld Meter MK09 Sound Lever Meter measurement device. We recorded noises of single propellers from different distances: $1$, $2$, $3$, $4$, and $5$\,m. In this test, we used a rigid propeller and Tombo propellers with configurations as shown in Table \ref{tab:Tombo configuration}.
%We employed four Conf. 3 propellers in take-off measurement, which is the noisiest working state of a drone, the results was summarised in Table \ref{tab: noise drone}.

\section{Results} \label{sec: results}
\begin{figure*}
	\centering
    \includegraphics[width=\textwidth]{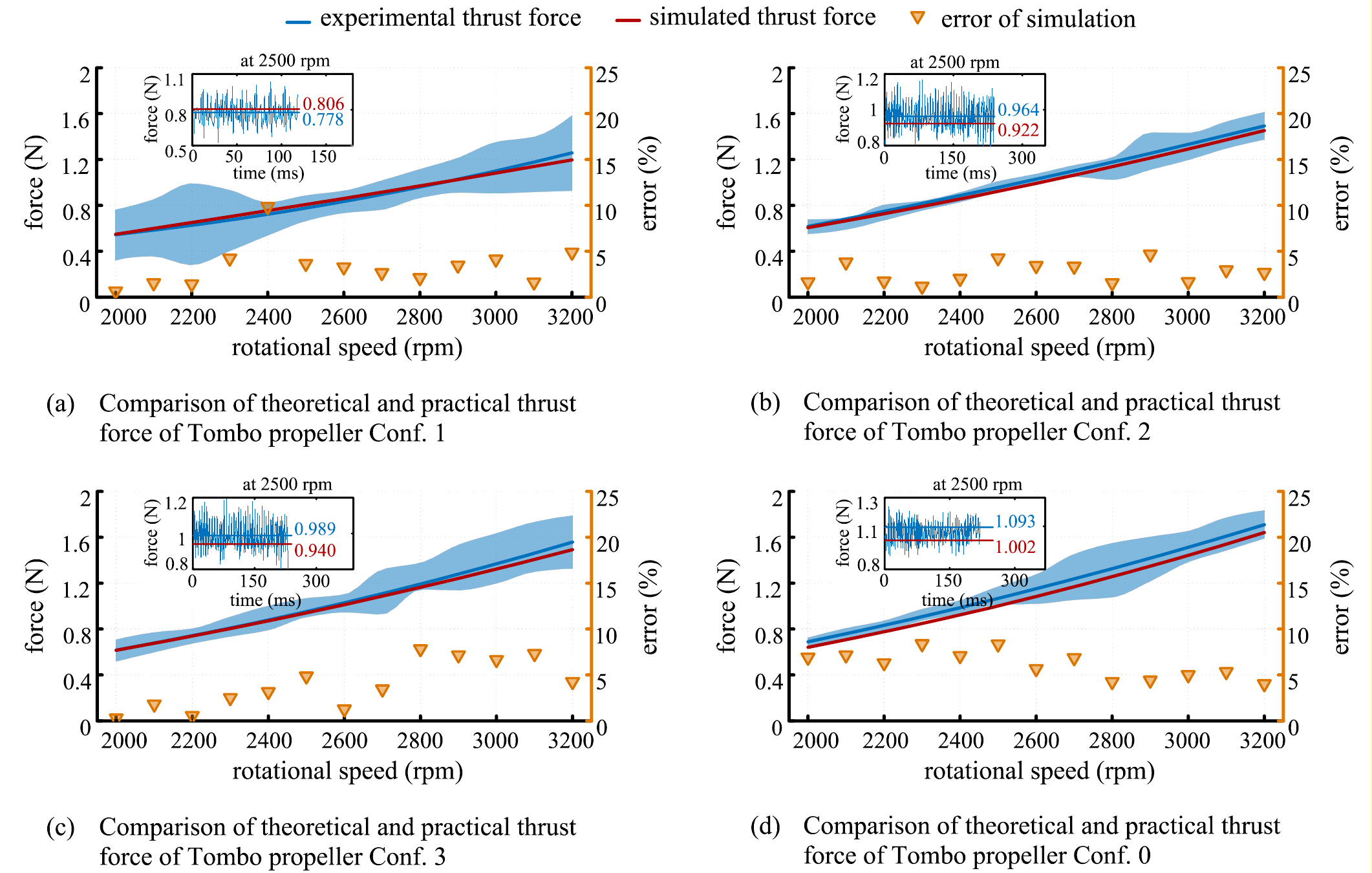}
    \caption{Comparison of thrust and collision response among Tombo propellers with different \textit{nodus} configurations. Experiments were conducted with three configurations of the Tombo propeller (Conf. 1, Conf. 2, and Conf. 3), and a rigid propeller (Conf. 0) in the speed range $2000$\,rpm to $3200$\,rpm. The red line plots the estimated thrust force (EsT), the blue line depicts the experimental thrust force (ExT), and yellow triangles indicate the error of simulation (EoS). The subgraphs (in boxes within graphs a, b, c, and d) show EsT and ExT of the propeller operating at a speed of $2500$\,rpm. (a) Conf. 1. (b) Conf. 2. (c) Conf. 3. (d) Conf. 0.}
	\label{fig: Evaluation of aerodynamics model}
\end{figure*}
\begin{figure*}
	\centering
    \includegraphics[width=\textwidth]{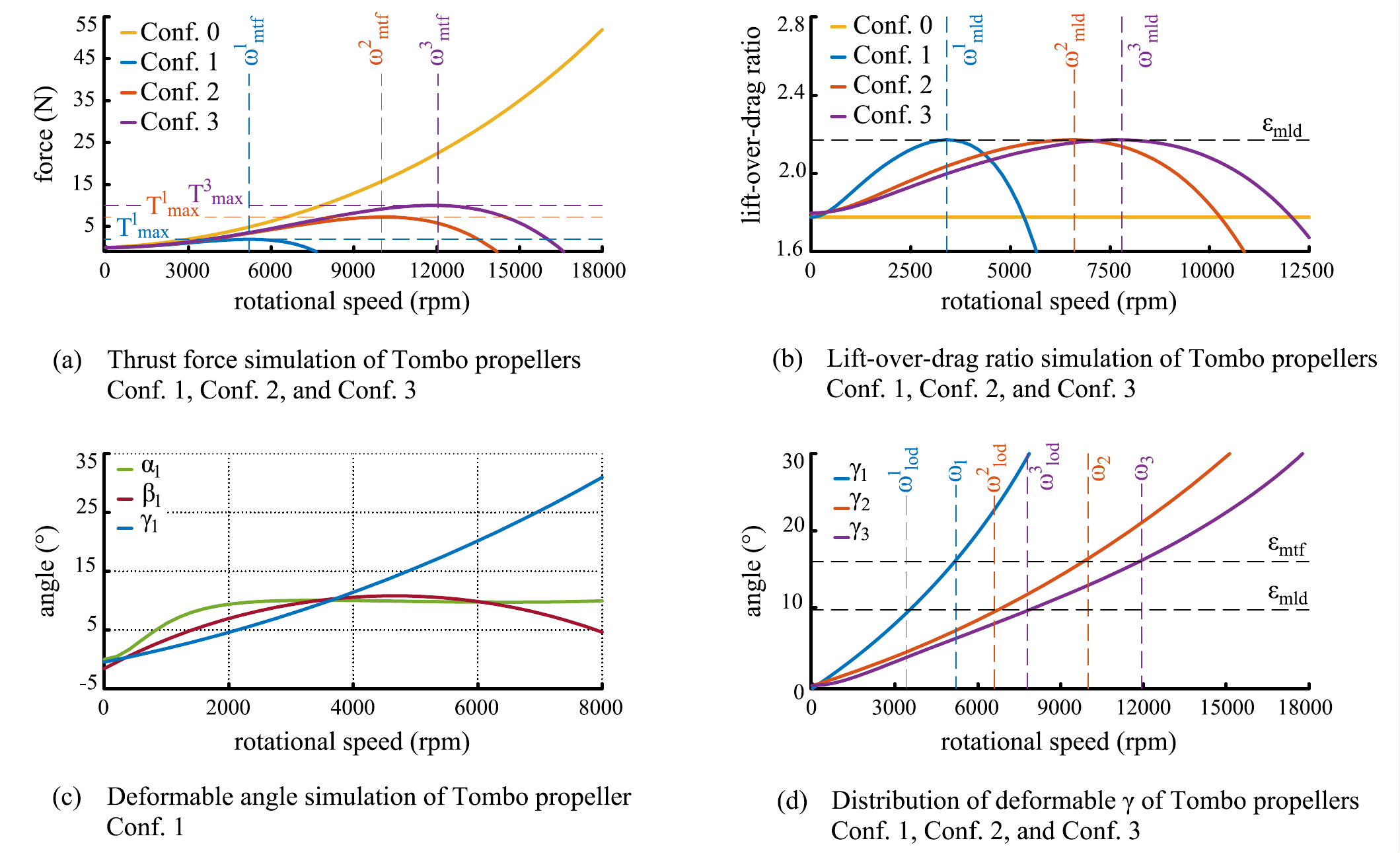}
    \caption{Aerodynamic parameter estimations of the Tombo propeller by \textit{nodus} stiffness. A rigid propeller (Conf. 0) and three Tombo propellers (Conf. 1, Conf. 2, and Conf. 3) were chosen for simulation. (a) We predicted thrust forces among the four configurations of propeller in a larger range of rotational speed (up to 18,000 rpm) to discover the distribution of these forces. $\omega_1$, $\omega_2$, and $\omega_3$ denote the rotational speeds of Conf. 1, Conf. 2, and Conf. 3, respectively when the thrust force reached the maximum value. (b) The lift-over-drag ratio of Conf. 1, Conf. 2, and Conf. 3 Tombo propellers changed with rotor speed and reached the maximum value $\varepsilon_{mld} = 2.17$ at rotational speeds $\omega^1_{lod}$, $\omega^2_{lod}$, and $\omega^3_{lod}$ respectively. (c) Three deformable angles of Conf. 1 versus rotational speed of rotor. (d) Twist angle $\gamma$ of Conf. 1, Conf. 2, and Conf. 3 shared the same value $\gamma_{mtf}$ and $\gamma_{mld}$ in both cases of maximum thrust force and maximum lift-over-drag ratio.}
	\label{fig: Estimations of aerodynamics model}
\end{figure*}
\subsection{Aerodynamic Model of the Tombo Propeller}
\subsubsection{\textit{Nodus}' Parameters}
The mechanical and geometrical properties of the \textit{nodus} are important parameters in construction of an aerodynamic model propeller. Note that the \textit{nodus} model presented in Section \ref{IV-C} requires the following parameters: length of the \textit{nodus}, representative cross-section information, number of fibers, diameter of fibers (from the design), Young modulus and shear modulus of the matrix material and fiber (from experiment results based on American Society for Testing and Materials Standard - ASTM 412 Die D dumbbell specimens with the Poisson ratio was chosen as $0.4$). Finally, Young modulus and shear modulus of the \textit{nodus} were calculated using Eq. \ref{eq: 2} and updated in Table \ref{tab:Tombo configuration}. The results indicated that the mechanical properties of the matrix play the most critical role in calculation of Young modulus and shear modulus of the nodus. In addition, increasing the number and diameter of the fibers enhanced these moduli. 
\subsubsection{Tombo Aerodynamics Model}
To evaluate the model, several experiments with different configurations of Tombo propeller (Table \ref{tab:Tombo configuration}) were conducted to compare estimated with  experimental aerodynamic parameters. Figure \ref{fig: Evaluation of aerodynamics model} shows evaluations of thrust force in the aerodynamic model for four different configurations of Tombo propeller. First, the model showed good thrust prediction when estimation errors were less than $8$\,$\%$ for different degrees of \textit{nodus} stiffness. Secondly, as \textit{nodus} stiffness increased magnitude and stability of lift force at the same speed increased. In the experimental rotational speed range (from $2000$ to $3200$\,rpm), the experimental and simulated thrust forces both exhibited linear characteristics, but the simulation results in a broader range (up to $8000$\,rpm) showed more clearly the nonlinear characteristics (see Figure \ref{fig: Estimations of aerodynamics model}a).
\par This model can help to simulate the aerodynamic parameters of the Tombo propeller over a more extensive range of rotational speed. Figure \ref{fig: Estimations of aerodynamics model}a shows the thrust force of rigid and Tombo propellers in the speed range below $18\,000$\,rpm. Both experimental and simulated results show that as the rotational speed gradually increased, the thrust force of Tombo propeller remained proportional to \textit{nodus} stiffness, and at higher speeds a significant difference was observed. However, unlike the rigid propeller, the thrust force of each Tombo propeller configuration reached maximum at critical thrust $T_{max}$ (in steady force state only). This critical thrust depends strongly on \textit{nodus} stiffness. The stiffer the \textit{nodus} the higher the critical thrust force generated. The rotational speed corresponding to this critical thrust is called the \textit{maximum thrust speed}, noted as $\omega_{mtf}$. When the rotational speed surpasses $\omega_{mtf}$, the generated thrust the generated thrust decreases rapidly. This phenomenon suggests that $\omega_{mtf}$ is a suitable choice for UAVs that need to carry heavy loads or when maximum acceleration is required.
\par Figure \ref{fig: Estimations of aerodynamics model}b shows the relationship between the lift-over-drag ratio $\varepsilon_{lod}$ and the rotational speed of the Tombo propeller, simulated from Eq. \ref{eq: 1} and \ref{eq: 5} with $\varepsilon_{lod} = F_l/F_d$. For rigid propellers, $\varepsilon_{lod}$ independent of the rotational speed because the propeller geometry does not change during rotation. For the Tombo propeller, the coefficient $\varepsilon_{lod}$ varied with rotational speed. However, as in the case of maximum thrust speed, each Tombo propeller has a \textit{critical lift-over-drag ratio rotational speed} $\omega_{mld}$ where $\varepsilon_{lod}$ reaches the maximum (denoted $\varepsilon_{mld}$). In other words, $\varepsilon_{mld}$ is independent of configuration of Tombo propeller being the same for all fabrication configurations. The reason is that these configurations all have the same initial geometric design, so when deformed, these configurations, even though they differ, will interfere in the state with the highest $\varepsilon_{lod}$ result. This critical lift-over-drag velocity $\omega_{mld}$ is variable between Tombo propellers and tends to increase with \textit{nodus} stiffness. Therefore, $\omega_{mld}$ can be chosen for high efficient work of UAVs such as traveling or delivery tasks. 
\par From Equation \ref{eq:alpha and beta} and \ref{eq: gamma} we can predict the deformable angles of the Tombo propeller. For instance, Figure \ref{fig: Estimations of aerodynamics model}c shows the simulation of these angles in a Conf. 1 propeller. It can be seen that as the propeller rotational speed increased, angles $\alpha$ and $\beta$ increased rapidly to reach a stable value of about $10$ degrees in the speed range from $2000$\,rpm to $5000$\, rpm. Meanwhile, $\gamma$ continued to increase rapidly showing no sign of saturation. Combined with Eq. \ref{eq: 8} and Figure \ref{fig: Estimations of aerodynamics model}a, this result demonstrated the significant contribution of $\gamma$ to the thrust force of the Tombo propeller. To confirm the simulated variation of these deformable angles, we experimented with angle $\beta$ using a Conf. 4 Tombo propeller (see Figure \ref{fig: nodusmodelling}f). This experiment showed simulated angle $\beta$ was similar to the experimental angle at a rotational speed range above $2000$\,rpm. In the lower speed range, the observed angle changed more strongly, due to a limitation of the experimental model (see Section \ref{subsec: deformableangle}). When the speed of the motor rotating the propeller was decelerated suddenly, the inertia force caused blade flapping at a rotational speed approaching zero, leading to a significant change in deformable angle.
\par To investigate the contribution of $\gamma$ to the aerodynamic of the Tombo propeller, we simulated this angle as shown in  Figure \ref{fig: nodusmodelling}f. Despite different configurations, Tombo propeller which share the same original design, will reach critical thrust and critical lift-over-drag ratio states at the same $\gamma_{mld}$ and $\gamma_{mtf}$. This finding strongly confirms that $\gamma$ plays the decisive role in the deformable states of the Tombo propeller.
\subsection{Characteristics of the Tombo Propeller with Different Configurations} \label{VI-B}
\begin{table*}
\caption{Characteristics of Tombo propellers and a rigid propeller measured at $2000$\,rpm of rotational speed}
\label{tab: survey of Tombo propellers}	
\centering
\begin{tabular}{lrrrrrr}
\hline
 &
  \multicolumn{1}{c}{\textbf{Thrust force {[}N{]}}} &
  \multicolumn{1}{c}{\textbf{\begin{tabular}[c]{@{}c@{}}Thrust force\\ deviation {[}N{]}\end{tabular}}} &
  \multicolumn{1}{c}{\textbf{Collision force {[}N{]}}} &
  \multicolumn{1}{c}{\textbf{Recovery time {[}s{]}}} &
  \multicolumn{1}{c}{\textbf{\begin{tabular}[c]{@{}c@{}}Simulated\\ L/D ratio\end{tabular}}} &
  \multicolumn{1}{c}{\textbf{\begin{tabular}[c]{@{}c@{}}Noise (distance: $5$\,m)\\ {[}dB{]}\end{tabular}}} \\ \hline
Conf.   0 &
  $0.658$ &
  $0.11$ &
  $269.3$ &
  -- &
  $1.776$ &
  $49.4$ \\
Conf.   5 &
  $0.656$ &
  \cellcolor[HTML]{99FF66}$0.03$ &
  $147.3$ &
  $0.63$ &
  \cellcolor[HTML]{99FF66}$2.0822$ &
  \cellcolor[HTML]{99FF66}$49.1$ \\
Conf.   6 &
  \cellcolor[HTML]{FFE699}$0.537$ &
  $0.14$ &
  $93.7$ &
  $0.55$ &
  $2.0718$ &
  $51$ \\
Conf.   7 &
  $0.631$ &
  $0.04$ &
  $81.5$ &
  $0.32$ &
  $2.0652$ &
  $49.6$ \\
Conf.   8 &
  $0.589$ &
  $0.09$ &
  \cellcolor[HTML]{99FF66}$80.4$ &
  $0.32$ &
  $1.9323$ &
  $48.7$ \\
Conf.   9 &
  $0.622$ &
  $0.07$ &
  $159.6$ &
  \cellcolor[HTML]{99FF66}$0.3$ &
  $1.9251$ &
  $49.2$ \\
Conf.   10 &
  $0.656$ &
  $0.12$ &
  $123.7$ &
  \cellcolor[HTML]{FFE699}$0.66$ &
  $1.9206$ &
  $50.2$ \\
Conf.   11 &
  \cellcolor[HTML]{99FF66}$0.666$ &
  $0.07$ &
  $123$ &
  $0.39$ &
  $1.9051$ &
  $49.5$ \\
Conf.   12 &
  $0.624$ &
  \cellcolor[HTML]{FFE699}$0.26$ &
  \cellcolor[HTML]{FFE699}$189.9$ &
  $0.55$ &
  $1.899$ &
  \cellcolor[HTML]{FFE699}$52.4$ \\
Conf.   13 &
  $0.611$ &
  $0.06$ &
  $145.5$ &
  $0.42$ &
  \cellcolor[HTML]{FFE699}$1.8952$ &
  $50.7$ \\ \hline
\textit{\textbf{Mean}} &
  \textit{\textbf{$0.621$}} &
  \textit{\textbf{$0.098$}} &
  \textit{\textbf{$127.2$}} &
  \textit{\textbf{$0.46$}} &
  \textit{\textbf{$1.9663$}} &
  \textit{\textbf{$50.0$}} \\ \hline
\end{tabular}
\end{table*}
\begin{figure*}
	\centering
    \includegraphics[width=\textwidth]{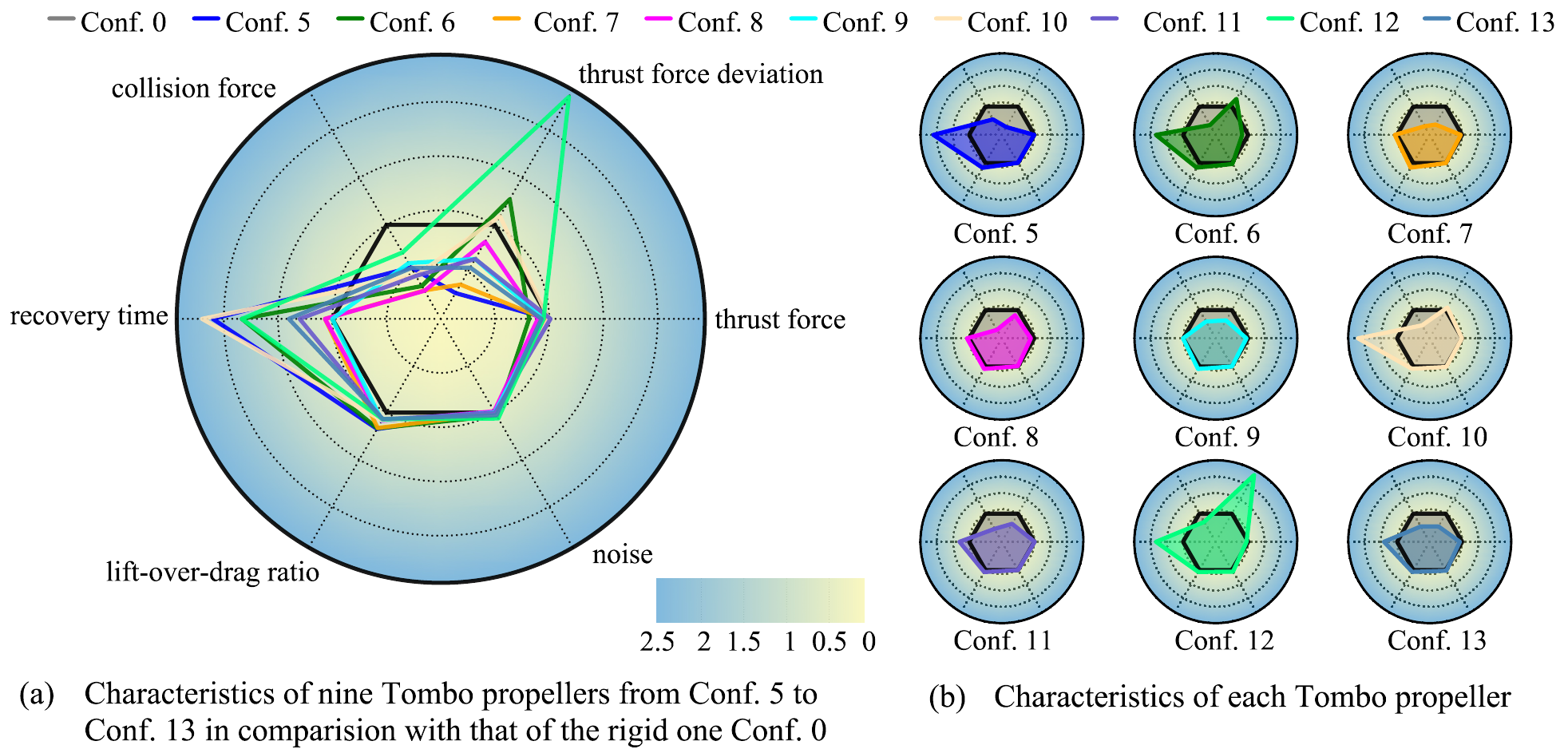}
	\caption{Comparison of characteristics among nine configurations of a Tombo propeller and a rigid propeller (normalized by metrics of the rigid propeller). (a) An overview of nine configurations (see Table \ref{tab:Tombo configuration}) of the Tombo propeller (colored lines) and a rigid propeller (black line) of the same morphology. (b) Metrics of each configuration Tombo propeller in comparison with the rigid propeller. Details of metrics can be found in Table \ref{tab: survey of Tombo propellers}.}
	\label{fig: tombo_parameters}
\end{figure*}
As mentioned above, the Tombo propeller is a novel propeller for UAVs, which was designed to contribute to ultimate safety. Therefore, it was important to investigate its characteristics to clarify and confirm its contributing features. Nine configurations of Tombo propeller and a rigid one were analyzed focusing on six characteristics: thrust force, thrust force deviation, collision force, recovery time, simulated lift-over-drag (L/D) ratio, and noise (Table \ref{tab: survey of Tombo propellers}). All experiments were conducted at a propeller rotation speed of $2000$\,rpm, and data were normalized and by the results of the rigid propeller (Conf. 0) for visualization as Figure \ref{fig: tombo_parameters}.
\par Overall, the Tombo deformable propeller shows promising characteristics compared to a rigid one. While all of thrust force, thrust force deviation, simulated L/D ratio, and noise of the experimental Tombo propellers ($0.621$\,N, $0.098$\,N, $1.9663$, and $50$\,dB, respectively) are almost the same as those of the rigid propeller ($0.658$\,N, $0.11$\,N, $1.776$, and $49.4$\,dB, respectively), thrust force deviation, collision force, and time recovery, all differed. Note that although the mean of thrust force deviation of the Tombo propeller was lower than that of the rigid propeller, its distribution was extensive, ranging from $0.03$ to $0.26$\,N among different configurations. Practical flights showed that thrust force deviation played a vital role in creating the balance of a drone, so the configuration needs to be considered to equip a drone. Last but not least, the collision force of the Tombo propeller was much smaller than that of the rigid propeller, resulting in lower risk of injury or damage upon collision with an obstacle in its surroundings.

\section{Tombo Propeller in Practical Flights and Its Responsiveness to Mid-air Collisions} \label{sec: flying experiment}
\begin{figure}
	\centering
    \includegraphics[width=\columnwidth]{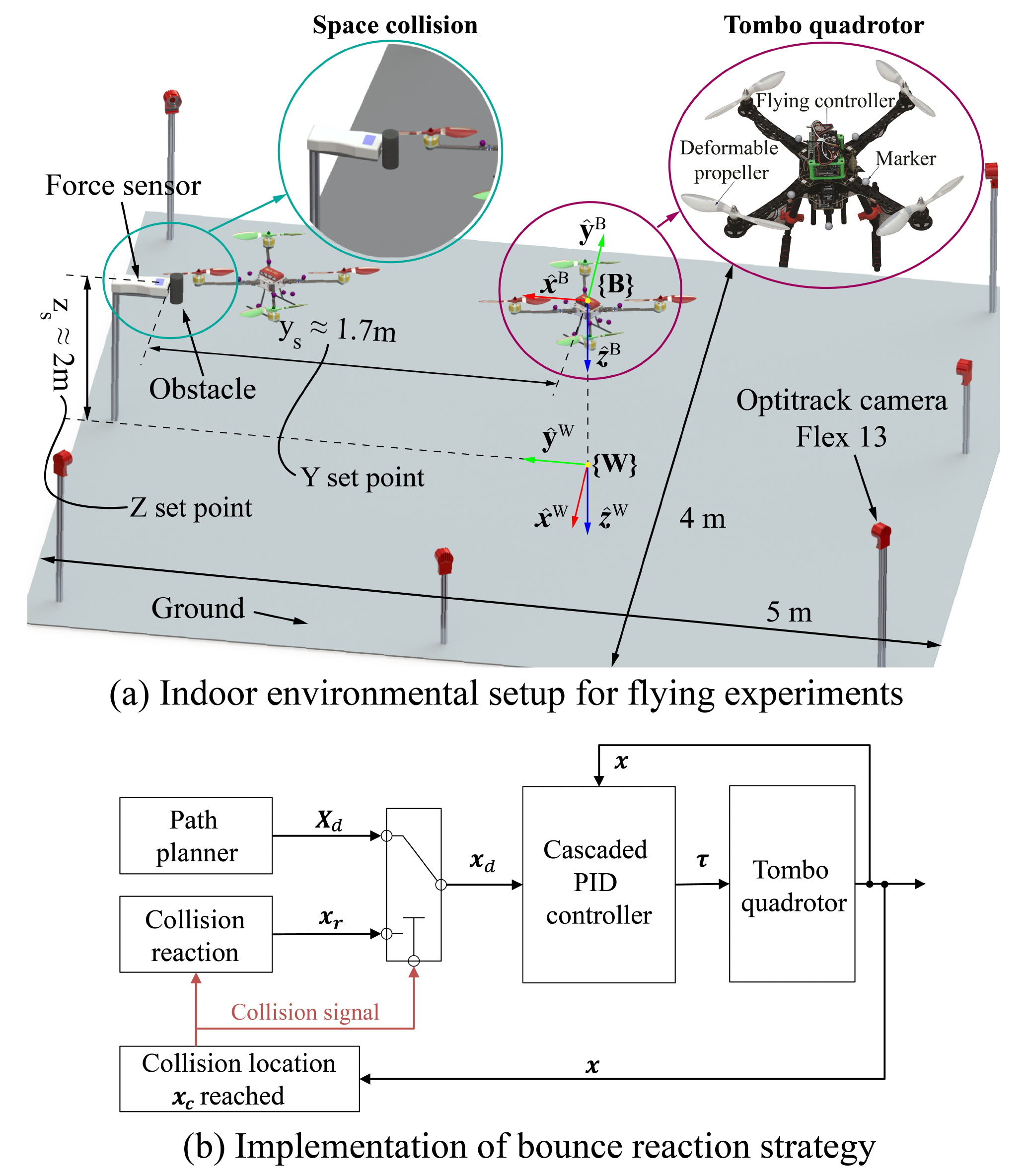}
	\caption{Experimental setup for drone flight experiments. (a) The indoor environment and setup for flight tests and collision experiments, in which the OptiTrack Mocap  system (for motion picture and 3D tracking) was used to determine the position the Tombo quadrotor. (b) Schematic of the reaction strategy which was implemented in response to a collision between the Tombo propeller and a fixed obstacle.}
	\label{fig: flying_experiment_setup}
\end{figure}
\begin{figure*}
	\centering
    \includegraphics[width=\textwidth]{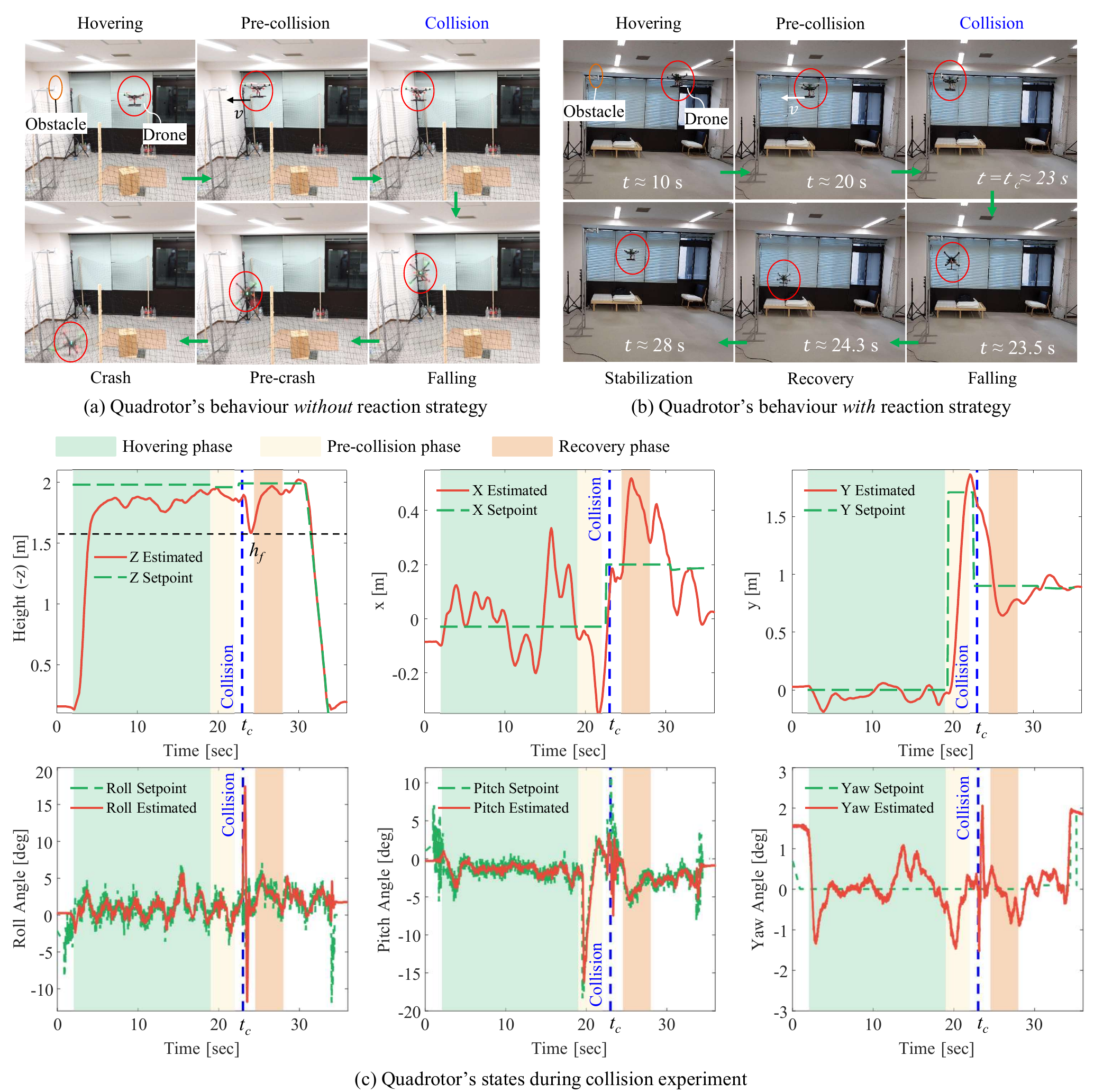}
	\caption{Examination of the equilibrium bounce reaction of a quadrotor upon a Tombo propeller-obstacle collision. (a) Without implementation of the reaction strategy, the quadrotor completely crashed to the ground after a collision. (b) Video stills demonstrate the effectiveness of the reaction strategy, which could stabilize the quadrotor within $5\,\text{seconds}$. (c) Logs of the quadrotor state in terms of position and orientation during the flight and collision experiment show at collision time $t_{c}$, the reaction control mode was triggered and attempted to stabilize the quadrotor at a safe position $\mathbf{x}_{r}=[0.2,\,0.9,\,-2.0]^\text{T}$\,m. (Note that the height of quadrotor is minus $z$-coordinate).}
	\label{fig: collision_recovery_results}
\end{figure*}
\par This section examines how the Tombo propellers behave in practical fight when equipped in drones (quadrotors, see Figure \ref{fig: flying_experiment_setup}a), as well as how the controller responds during and after a collision between the Tombo drone and a fixed mid-air obstacle. We chose Tombo propellers Conf. 13 to conduct the experiments in this section because they have a low thrust deviation (see Section \ref{VI-B}) and the highest stiffness of the $nodus$ and the deformable edge (see Table \ref{tab:Tombo configuration}), resulting in ease of fabrication and high consistency among fabricated propellers. Additionally, to leverage the self-recovery ability of our deformable propellers, we implemented an equilibrium bounce reaction scheme \cite{Haddadin} which was specifically tailored to rescue the Tombo drone (quadrotor) from a sudden fall in event of propeller-obstacle collision.
\subsection{Reaction Strategy}
In pursuit of drone safety provided by deformable propeller recovery, we investigated an equilibrium bounce reaction strategy expected to prevent the Tombo quadrotor from falling in event of a propeller-obstacle collision. We considered a case where the quadrotor flew at designated trajectory $\textbf{X}_{d}$ such that one of the front propellers collided with a mid-air obstacle located at  designated position $\mathbf{x}_{c}\in\mathbb{R}^{3}$ with reference to the global $\{W\}$ coordinate frame (Figure \ref{fig: flying_experiment_setup}a). Once $\mathbf{x}_{c}$ reached; thus collision occurred, the reaction mode was immediately triggered to set the equilibrium position to $\mathbf{x}_{r}=\mathbf{x}_{c}+d_{r}\mathbf{n}_{r}$, where $d_{r}>0$ is the bounce distance, and $\mathbf{n}_{r}$ is the reactive normal that is opposite to the flying direction before the time of collision. This encourages the low-level controller to produce virtual forces that drive the quadrotor toward equilibrium state $\mathbf{x}_{r}$, and then enable stabilization of the vehicle at a safe distance from the obstacle. Since the falling rate of the quadrotor, as observed in experiments, was determined to be around $0.3\,$m/s, and the propeller recovery time was determined to be approximately $0.42\,$seconds, a typical PID (proportional–integral–derivative) cascaded controller could be a possible solution for low-level control. The cascaded control architecture of the quadrotor includes an outer loop P position, followed by the P angle and PID angular rate controllers, running at $50\,$Hz, $250\,$Hz and $1\,$kHz, respectively. The equilibrium bounce reaction strategy is summarized in Figure \ref{fig: flying_experiment_setup}b and  and we open-source the implementation of the reactive control strategy using ROS (Robot Operating System) \cite{ros} as follows \url{https://github.com/Ho-lab-jaist/tombo-propeller.git}.

\subsection{Experimental Setup}
Figure \ref{fig: flying_experiment_setup}a illustrates the indoor environment where the flight and collision experiments were conducted. Inside the experimental space, the custom-built quadrotor, mounted with 8 reflective markers, was precisely positioned by the \emph{OptiTrack} motion capture system (mocap) including 6 tracking cameras (OptiTrack Flex 13), which provide an overall positional accuracy up to $1.2\,\text{mm}$. The quadrotor's 6-DoF pose (\emph{i.e.}, position and orientation), estimated by the mocap system on a desktop PC (Intel i7-7700\,CPU at 3.6GHz and 8\,GB RAM), was communicated to an onboard computer (Jetson TX2, NVIDIA) at around $30\,\text{Hz}$ so that the automatic flight and collision reaction scheme can be practically implemented in real time, with the assistance of the PID-based low-level controller (PX4 flight stack\footnote{https://px4.io/}) running on the \emph{Pixhawk 4} autopilot hardware. While the real-time postural information provided feedback signals for the P position and angle controllers, the onboard inertial measurement unit (IMU) was responsible for the most inner PID control loop of angular rate.

\subsection{Results: Flying Behavior, Collision Response and Recovery} \label{sec: result_collision_recovery}
To investigate the responsiveness of the Tombo propeller and effectiveness of the bounce reaction (recovery process) scheme upon the aforementioned collision, we set up a flight test wherein the quadrotor flew along planned trajectory $\textbf{X}_{d}$ in $\hat{y}$ direction such that the front port (left) propeller collided with an obstacle located at position $\mathbf{x}_{c}=[0.0,\,1.7,\,-2.0]^\text{T}$\,m. Upon the collision, we set the bounce distance $d_{r}=0.8\,$m, leading to the equilibrium position at $\mathbf{x}_{r}=[0.2,\,0.9,\,-2.0]^\text{T}$\,m, for activation of the recovery process. The Tombo quadrotor, as observed from the experiments (see Figure \ref{fig: collision_recovery_results}), could achieve stable flights as it can track the specified reference trajectory of position $\textbf{X}_{d}$ during the hovering and pre-collision (flying along the $y$-direction) phases (Figure \ref{fig: collision_recovery_results}c). The quadrotor started to fall when a propeller collided with the obstacle. Without application of the reaction scheme, the drone could not recover and would fall and crash (as shown in Figure \ref{fig: collision_recovery_results}a). In contrast, application of the bounce reaction strategy, in addition to the fast recovery of the Tombo propeller, allowed the quadrotor to stabilize in response to the collision (see Figure \ref{fig: collision_recovery_results}b). Figure \ref{fig: collision_recovery_results}c details the behavior of the quadrotor during the collision and recovery process. Specifically, once the collision occurred at around $t_{c}=23\,\text{seconds}$, the quadrotor oscillated and overturned with a large fluctuation in roll angle, then started to fall, which characterizes the unstable phase. However, the quadrotor did not fall to the ground (collision height 2\,m). It took about $1.5\,\text{seconds}$ to overcome the unstable phase, followed by a recovery process of about $3.5\,\text{seconds}$ before attaining a stable state, which confirmed the responsiveness of the Tombo propeller and the effectiveness of the reaction strategy in event of collision with an obstacle. A video demonstration of the collision experiment is at: \url{https://youtu.be/qKqtqpq_yqw}.   

Consequently, in this section, we confirmed the flight ability of the drone with Tombo propeller in a real platform and the recovery ability after the collision. The obtained results reveal that with minimal invasion of the classical control strategy, the drone with Tombo propeller still can perform basic flight/hovering and novel reaction upon collision with the surrounding. As a result, introduction of softness to the propeller not only decreases the risk of damage, but also does not necessarily compromise the flight ability of the drone.

\section{Discussion} \label{sec: discussion}
\subsection{Fabrication}
Integration of soft materials into a conventional propeller improved safety in the operation of drones, especially by recovery ability and risk mitigation upon collision. For mass production of the Tombo propeller, the fabrication process proposed in this research needs to be simplified. The process of embedding tendons into the matrix by gluing them into the rigid parts (hub and wing) required significant manual dexterity. This, in fact, affects the quality of the fabricated propeller, thus decreasing its durability in long-term usage. Note that, during implementation of recovery trials ($10$ trials) in the previous section, we needed to replace the collided propeller three times, averagely one Tombo propeller may endure up to three collisions. Regarding the scalability, we have succeeded in fabrication of a wide range of Tombo propellers, such as $5$, $9$, $10$, $20$-inch long propellers, based on the proposed design and fabrication method. Smaller or bigger size and shape of the propeller will be investigated in the future.

Through evaluation experiments, we found that during rotation the entire Tombo propeller from the hub to the tip became quite stiff due to centrifugal force. Therefore, if connection between the \textit{nodus} and the rigid parts is assured, the tendons used in the present design may be made redundant. In that event, the double injection method may be exploited for mass production of the Tombo propeller, since this technique can create a dependable junction between the soft and hard parts, as well as the longevity of the propeller. More elaboration on this approach will be conducted in future work.
\subsection{Flight Ability}
The results obtained in experiments revealed, for the first time, successful flight performance and collision recovery behavior of a drone with a deformable propeller. However, there were some limitations in this study. The experiment trials were performed indoors, which ignores outdoor factors such as wind and weather conditions. Also, although PID controllers might achieve better performance in rigid propellers, due to the shortcomings of typical PID controllers, tracking errors in normal flight with deformable Tombo propellers were relatively large in some cases, especially in the $\hat{x}$-direction (Figure \ref{fig: collision_recovery_results}c), which led to different collision directions among flight trials and to various recovery behaviors (shown in the video at the link above). Also the fabricated propellers might not have yielded consistent behavior, which may have affected the controller's operation. In fact, the results shown in this article present the most typical drone behavior in response to the critical case of collision and to provide a benchmark against reactive performance. The average falling recovery time $\Delta t_{\text{recovery}}=5\,\text{seconds}$, and the averaged maximum falling distance $\Delta h_{\text{fall}}=0.5\,\text{m}$, were acceptable for this control strategy, considering the mid-air collision height was 2\,m indoors. For higher altitudes there would more time thus a better chance for the drone to fully recover before crashing to the ground. That leaves rooms for work in future improvement of reaction control algorithms (\emph{e.g.}, model predictive control, impedance/admittance reactive controllers \cite{Haddadin}) since the recovery time of the tested propellers ($\approx0.46\,$seconds) was shorter than the current time for falling recovery ($\approx5\,$seconds). To this end, in future work we aim to construct a comprehensive dynamics model of quadrotors that takes the aerodynamic modeling of Tombo propellers into consideration, which forms the foundation for advanced model-based interaction and tracking controllers, and also for morphological design optimization. Moreover, integration of perception to detect and avoid potential collisions is crucial to the realization of fully autonomous agile and resilient flying robots, which will be considered in future work.

\subsection{Possible Applications}
The Tombo propellers are expected to be loaded on drones for reducing damage risk of obstacle-propeller collision in any direction. It also brings in the recovery chance after collision, thus mitigating risk to the drone itself and properties on the ground compared to crashing to the ground. The use of the Tombo propellers with other measurements will ultimately increase the safety of drones in tasks close to objects or humans (infra inspection, freight, and so on). We  would  also  like  to  adopt  this biomimetic  approach  for  applications  in  other  fields  such  as small-scaled wind power generation propellers (reducing bird-strike risk) or ship propellers (reducing entanglement with marine litter, fish-strike risk) toward sustainable solution for the nature.      

\section{Conclusion} \label{sec: conclusion}
In this work, we proposed the design and fabrication of a biomimetic propeller. This approach can be applied to different types of propellers (\textit{e.g.} flapping wing and glider wings). The proposed aerodynamic model showed to correctly estimate the propeller parameters including thrust force and deformation angle in the plane perpendicular to the rotor plane. An examination of the characteristics of the Tombo propeller clarified the features and practicality of using this novel design for different vehicles. In addition, multiple flight experiments also demonstrated the ability of the Tombo propeller to increase drones' resilience to collisions, while concurrently preserving its mechanical structure after the impact.
In the future, we aim to enhance the aerodynamic model of the Tombo propeller taking into account the contribution of the deformable edge. Furthermore, we would like to develop a software application leveraging this aerodynamic model to automatically produce aerodynamic metrics and automatically recommend a biomimetic design for a conventional propeller as an output to the user.

\ifCLASSOPTIONcaptionsoff
  \newpage
\fi

\section*{Acknowledgments}
This work was supported by JST SCORE project, Grant-in-aid for Scientific Research projects No. 18H01406 and 21H01287. We also thank Dr. Pho Van Nguyen for his advice on the modeling approach, Dr. David Price for his thorough proofreading of this manuscript.

\bibliographystyle{IEEEtran}
\bibliography{main}

% \begin{IEEEbiography}{Son Tien Bui}
% Biography text here.
% \end{IEEEbiography}

% % if you will not have a photo at all:
% \begin{IEEEbiographynophoto}{Quan Khanh Luu}
% Biography text here.
% \end{IEEEbiographynophoto}

% \begin{IEEEbiographynophoto}{Dinh Quang Nguyen}
% Biography text here.
% \end{IEEEbiographynophoto}

% \begin{IEEEbiographynophoto}{Nhat Dinh Minh Le}
% Biography text here.
% \end{IEEEbiographynophoto}

% \begin{IEEEbiographynophoto}{Giuseppe Loianno}
% Biography text here.
% \end{IEEEbiographynophoto}

% \begin{IEEEbiographynophoto}{Van Anh Ho}
% Biography text here.
% \end{IEEEbiographynophoto}

\end{document}